\documentclass[Afour,sageh,times]{sagej}
\usepackage{sagej_arxiv} 

\usepackage{url}
\usepackage{enumitem}
\usepackage{xcolor}
\definecolor{ijrrblue}{RGB}{0, 51, 153} 

\usepackage{siunitx}
\DeclareSIUnit{\fps}{fps} 
\usepackage[
    colorlinks=true,
    bookmarksopen=true,
    bookmarksnumbered=true,
    citecolor=ijrrblue,
    linkcolor=ijrrblue,
    urlcolor=ijrrblue
]{hyperref}

\newif\ifshowvoiceover
\showvoiceoverfalse 

\usepackage[printonlyused]{acronym}

\newcommand\BibTeX{{\rmfamily B\kern-.05em \textsc{i\kern-.025em b}\kern-.08em
T\kern-.1667em\lower.7ex\hbox{E}\kern-.125emX}}

\def\volumeyear{2016}

\begin{document}

\runninghead{Jenelten and Ishtaweera et al.}

\title{Autonomous Trail Racing with a Bicycle Robot}

\author{%
Fabian~Jenelten\affilnum{1,$\dagger$},
Deepana~Ishtaweera\affilnum{1,2,$\dagger$},
Pablo~Garcia\affilnum{1},
Liana~Margolese\affilnum{1},
Igor~Bogoslavskyi\affilnum{1},
Luca~Buonocore\affilnum{1},
Jan~Migliazza\affilnum{1},
Aleksandrs~Ecins\affilnum{1},
Annika~Wollschlaeger\affilnum{1},
Dario~Bellicoso\affilnum{1},
Gabriel~Nelson\affilnum{3},
Abel~Gawel\affilnum{1} and
Marco~Hutter\affilnum{1,2}
}

\affiliation{\affilnum{1}Robotics and AI Institute (RAI), Zurich, Switzerland.\\
\affilnum{2}Robotic Systems Lab, ETH Zurich, Zurich, Switzerland.\\
\affilnum{3}Robotics and AI Institute (RAI), Cambridge, MA, USA.\\
\affilnum{$\dagger$} These authors contributed equally to this work.
}

\corrauth{Fabian Jenelten and Deepana Ishtaweera, 
Robotics and AI Institute (RAI), Zurich, Switzerland; 
Robotic Systems Lab, ETH Zurich, Zurich, Switzerland.}
\email{fjenelten@rai-inst.com, dishtaweera@rai-inst.com}

\begin{abstract}
Navigating off-road trails on a mountain bike requires diverse locomotion skills, including aerial maneuvers, cornering, and slip recovery, paired with sophisticated perception for scene understanding. Translating these capabilities to autonomous, two-wheeled robots remains a formidable challenge. 
We show that autonomous, high-speed trail navigation can be achieved through a single idea: encoding the trail as a semantic layer of the terrain. In simulation, this layer embeds the navigation task into the training environment; during deployment, it is reconstructed onboard from camera and LiDAR data. 
First, we introduce a unified learning framework that treats local navigation and locomotion as a joint optimization problem, bypassing the twist or pose command interface of classical control hierarchies. With fixed reward weights, the same framework produces both trail racing and agile parkour, shaped only through terrain semantics and episode length.
Second, our perception pipeline combines a frozen vision foundation model with a lightweight dual-head decoder that predicts per-pixel terrain semantics and confidence scores, which are fused with LiDAR range measurements into an egocentric semantic elevation map at \SI{30}{\hertz} onboard an NVIDIA AGX Orin.
We validate our framework on the \ac{UMV}, a bicycle robot that autonomously raced across all sections of a kilometer-long, unstructured mountain bike trail at speeds exceeding \SI{5}{\meter\per\second} and averaging \SI{3.26}{\meter\per\second}. In simulation, our policy fails up to an order of magnitude less often than hierarchical baselines while driving faster.
We release our simulation and perception tools as open source.
\acresetall
\end{abstract}


\maketitle

\section{Introduction} \label{sec:introduction}
\begin{figure}[h!]
\centering
\includegraphics[width=\columnwidth]{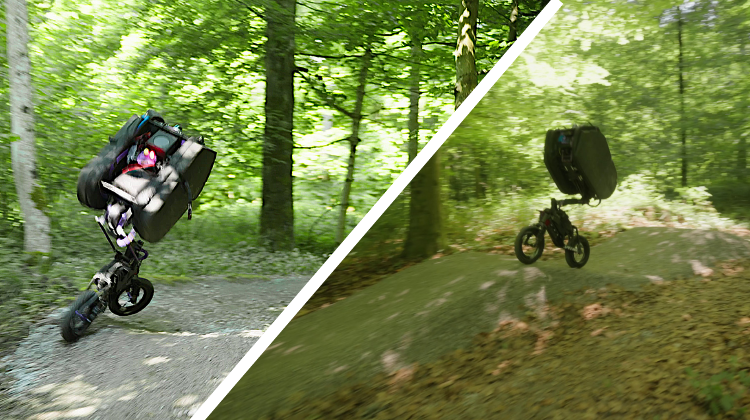}
\caption{\textbf{The \ac{UMV} autonomously navigates a mountain-bike trail in Zurich.} Left: The robot corners a curve. Right: The robot traverses pump-track waves.}
\label{fig:main}
\end{figure}

Off-road mountain biking demands a robust fusion of high-level world understanding, spatial planning, and split-second physical coordination. Deploying statically unstable systems, like a bicycle robot, on a mountain bike trail is inherently difficult because uneven, unpredictable terrain in outdoor forests makes it hard to plan and execute maneuvers. Friction fluctuates with changing ground conditions, seasonal shifts alter vegetation, and changing sun positions and cloud coverage introduce unpredictable shadows and harsh lighting variations. Therefore, \emph{autonomous trail racing} emerges as a profoundly challenging, tightly coupled perception-and-control problem. It combines research challenges from navigation~\citep{ravichandran2022memory, rana2023sayplan, lee2024wheeled, werby2024hovsg, rajvanshi2024saynav}, autonomous hiking~\citep{miki2023bridging, mattamala2025wvn, Reed2025, Giusti2016trail}, legged parkour~\citep{hoeller2024parkour, cheng2024extreme, kim2025highspeed}, high-speed autonomous ground vehicle racing~\citep{kabzan2020amz, cataffo2022nmpc, betz2023tum}, and drone racing~\citep{Loquercio2021droneracing, kaufmann2023swift, bahnam2026monorace}, while adding the complexity related to the unstable bike dynamics. Autonomous trail racing sits at the intersection of these domains, here summarized as ``\emph{time-optimal traversal of mountain bike trails relying strictly on onboard compute and sensing using wheeled, statically unstable platforms}''. Each of the aforementioned fields addresses one or more relevant aspects, but no method provides a complete solution to trail racing. We tackle this challenge by (i) unifying local navigation and control and (ii) casting pathfinding as an explicit perception problem, without the need for a global map or destination-based planning.
\begin{figure*}[h!]
\centering
\includegraphics[width=\textwidth]{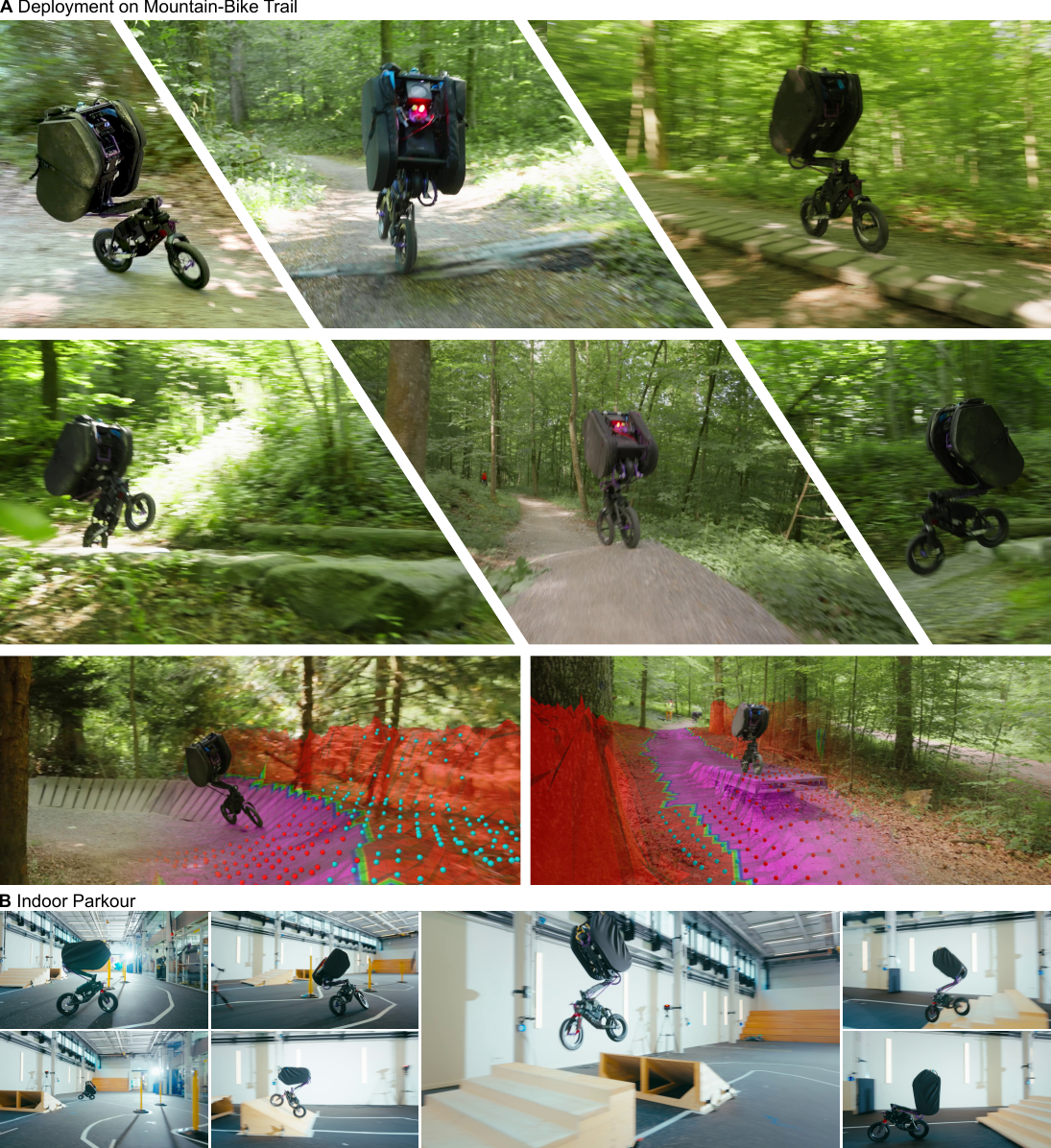}
\caption{\textbf{Autonomous trail racing and dynamic parkour.} \textbf{(A)} The \ac{UMV} autonomously navigates a mountain bike trail in Zurich at speeds exceeding \SI{5}{\meter\per\second}, relying exclusively on onboard perception and computation. The robot masters diverse obstacles, including roots, rocks, waves, and tabletop drops (first and second row). The policy jointly combines control and local navigation, while the two-layered grid map encodes a traversable path (third row). \textbf{(B)} With only the episode length and mesh as tuning degrees, the identical learning pipeline generalizes to agile parkour: A single policy navigates a slalom segment, accelerates up a ramp at \SI{5}{\meter\per\second}, clears a \SI{1.2}{\meter} long gap, lands on a five-step staircase, and executes a track-stand, all within an \SI{8}{\second} window.}
\label{fig:teaser}
\end{figure*}

\begin{figure*}[h!]
\centering
\includegraphics[width=\textwidth]{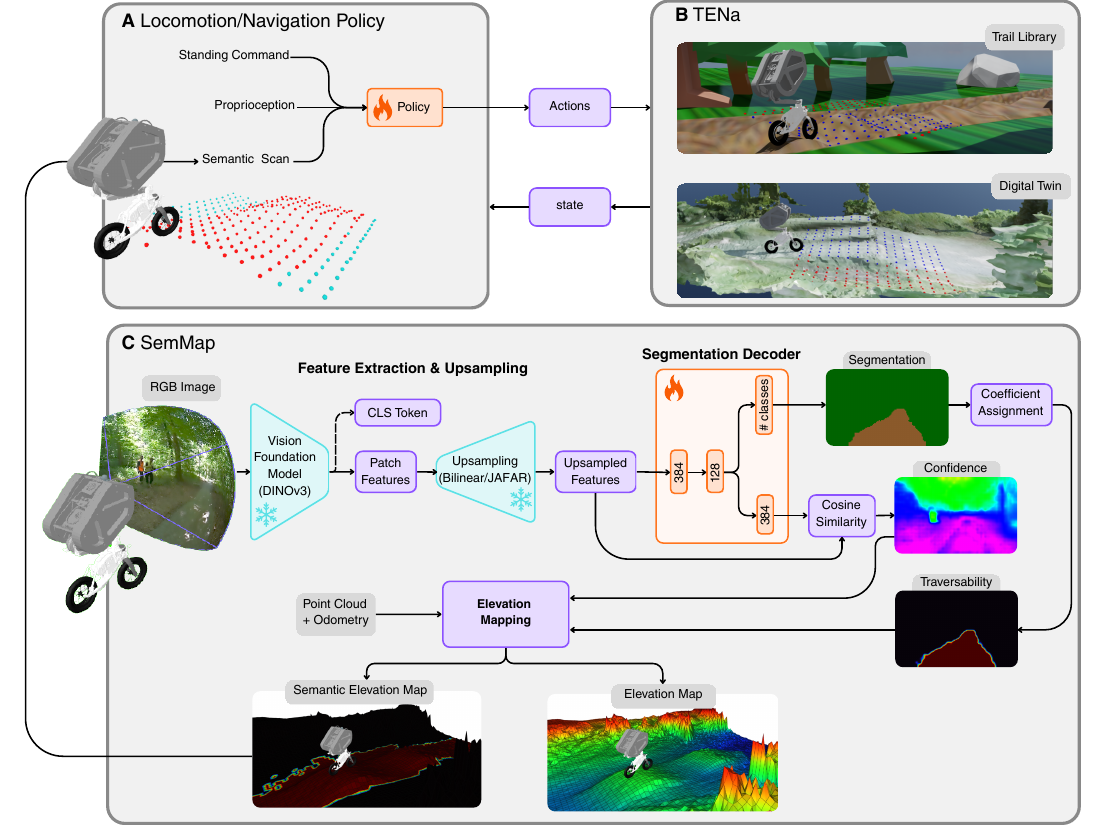}
\caption{\textbf{Approach overview.} \textbf{(A)} \emph{Deployment:} The policy, parametrized as \ac{MLP} with three hidden layers, observes proprioceptive states, an operator-triggered stand command, and a semantic height scan. Joint action targets are computed at~\SI{50}{\hertz} to control the rear wheel velocity, the fork angle, and the two \acp{DOF} of the torso. \textbf{(B)} \emph{Training:} Our learning framework utilizes synthetic trails or high-fidelity digital twins and applies to both trail racing and parkour. \textbf{(C)} \emph{Semantic pipeline}: An incoming RGB image passes through a frozen DINOv3 model and a frozen upsampler to produce high-resolution per-patch features. A lightweight trainable decoder maps these features to a per-pixel segmentation mask. A parallel reconstruction head, scored by cosine similarity to the original features, yields a per-pixel confidence score. A coefficient assignment operation converts the segmentation mask into a traversability image. Finally, an elevation mapping pipeline fuses traversability and confidence scores with odometry and point clouds at~\SI{30}{\hertz}.}
\label{fig:approach}
\end{figure*}

The prior work by~\cite{bokser2026umv} presented the \ac{UMV}, a two-wheeled bicycle robot capable of high-speed driving up to \SI{8}{\meter\per\second} on flat ground. Two actuated \acp{DOF} in the upper body allow the robot to trigger vertical jumps reaching \SI{1}{\meter} in height. While it demonstrated the physical feasibility of driving down a trail within a human-operated experiment, we consider the \ac{UMV} the ideal platform to push toward full autonomy and high-speed racing. Expanding upon this hardware foundation, we integrate onboard perception, develop real-time visual world understanding, and learn vision-locomotive skills to present a comprehensive framework for autonomous trail racing (Fig.~\ref{fig:approach} and multimedia extensions~\ref{mov:method} and~\ref{mov:allmend_racing}).

We break this challenge down into two core scientific contributions. First, we introduce a platform-agnostic \ac{MDP}, herein referred to as \ac{TENa} (subsection~\ref{ssec:tean}). By encoding the traversable path as a semantic layer directly into the terrain mesh and treating it as a soft boundary during training, locomotion and local navigation collapse into a unified optimization problem. To support our specific application, we also present the \ac{TL}, an approach for the procedural generation of diverse, unidirectional synthetic trails that enables zero-shot deployment of racing policies (subsection~\ref{ss:terrain_gen}). Second, to render the semantic path observable to the control policy during real-world deployment, we introduce \ac{SemMap} (subsection~\ref{ss:semmap_method}). This module constructs a two-layered, robot-centric elevation map containing both geometric and semantic path information. Leveraging a frozen \ac{VFM} paired with a lightweight decoder, incoming RGB images are processed at up to \SI{90}{\hertz} in isolation and fused with LiDAR range measurements at \SI{30}{\hertz}, producing a 2.5D elevation map that holds the path as a segmented trail.

We demonstrate autonomous deployment on a kilometer-scale, real-world mountain bike trail, sustaining average speeds of \SI{3.26}{\meter\per\second} and peaking above \SI{5}{\meter\per\second} (Fig.~\ref{fig:teaser}A). Compared to prior work~\cite{bokser2026umv}, our work (i) introduces a single training environment applicable to both trail racing (subsections~\ref{ssec:trail_indoor}, \ref{ssec:trail_racing}) and parkour (subsection~\ref{ssec:parkour}) with fixed reward weights; (ii) removes the human operator by following a trail observed by~\ac{SemMap} (subsections~\ref{ssec:semmap}, \ref{ssec:trail_indoor}); (iii) outperforms the teleoperated policy by a factor of 2.3 in speed on a real-world mountain-bike trail where baseline data is available (subsection~\ref{ssec:trail_racing}); (iv) in simulation, fails up to an order of magnitude less often than hierarchical baselines built on the policy of \cite{bokser2026umv}, including one with a learned navigation policy, while driving faster  (subsection~\ref{ssec:comparison_hierarchical}); and (v) integrates the necessary state-estimation tools and onboard perception hardware (subsection~\ref{ssec:deployment_hardware}). Compared to related works in autonomous hiking, we (i) remove the command interface by learning locomotion and local navigation as a joint optimization problem to achieve high-speed traversal; and (ii) demonstrate these capabilities on a statically unstable bicycle robot. By open-sourcing our core modules, \ac{TL} and \ac{SemMap}, this work establishes a comprehensive baseline for autonomous trail racing.\endnote{\ac{TL} (\url{https://github.com/isaac-sim/IsaacLab/pull/6539}) and \ac{SemMap} (\url{https://github.com/rai-opensource/sem_map}).}

\section{Related Work} \label{sec:related_work}
In robotics, the problem of \emph{navigation} -- defined as moving from an initial state to a target location -- is typically solved by separating planning from control. This hierarchy relies on high-level representations such as trajectories, waypoints, checkpoints, or language. Most often, a \emph{global navigation} planner acts on a pre-generated~\citep{ravichandran2022memory, rana2023sayplan, lee2024wheeled, werby2024hovsg, miki2023bridging} or online-built graph~\citep{shah2023vint, rajvanshi2024saynav}. A \emph{local navigation} planner then follows this global path while accommodating additional constraints, such as obstacle avoidance, before a downstream locomotion controller executes the high-level commands. To achieve higher overall performance and agility, locomotion and local navigation can be jointly solved by tracking pose targets~\citep{Loquercio2021droneracing, rudin2022advanced, hoeller2024parkour}.  Our work follows this trend by fully absorbing local navigation into the locomotion task, which, in simulation, reduces failures by up to an order of magnitude compared to hierarchical baselines (subsection~\ref{ssec:comparison_hierarchical}).

The problem of \emph{racing} -- defined as minimizing lap time while adhering to a predefined path -- has attracted considerable research, leading to a variety of control methods deployed across diverse robotic platforms. In autonomous \emph{ground-vehicle racing}, the environment typically simplifies to a 2D track. Here, most approaches employ \ac{MPC}~\citep{goldfain2019autorally} or follow a model-based planning-and-tracking paradigm~\citep{srinivasan2021holistic}. For example, \cite{liniger2015optimization} introduced a mathematical optimization framework using a nonlinear \ac{MPCC} controller for RC cars, a method later adapted for small-scale Formula-style racing cars~\citep{kabzan2020amz}. Leveraging offline optimization combined with online \ac{MPC}~\citep{betz2023tum} enabled a full-scale autonomous vehicle to win the Indy Autonomous Challenge. While end-to-end \ac{RL}~\citep{fuchs2021superhuman, song2021autonomous, zhang2021roach, Stachowicz2023FastRLAP} offers potential deployment advantages, classical optimization-based methods continue to dominate ground-vehicle racing. This dominance is likely driven by their ability to provide strict mathematical guarantees for safety-critical spatial constraints.

In \emph{aerial drone racing}, the 2D track-following problem expands into a 3D gate-traversal task. Early approaches tackled the mathematical challenge of generating time-optimal trajectories through multiple waypoints offline~\citep{foehn2021time}. Later methods introduced online re-planning and gate-collision constraint satisfaction using a variant of \ac{MPCC}~\citep{krinner2024mpcc}. Laying the groundwork for physical deployment, \cite{foehn2022alphapilot} detailed a comprehensive, modular autonomous racing system, establishing the architectural blueprint for modern drone racing pipelines. Building upon this vision backbone and its gate detector, recent work demonstrated that a learned control policy can outperform human expert pilots by directly observing the next gate location~\citep{kaufmann2023swift}. As noted by~\cite{song2023reaching}, the growing dominance of \ac{RL} in this field is driven by its ability to directly optimize for the holistic task objective, coupled with increased robustness achieved through domain randomization. Given a predefined and observable track, our work draws direct inspiration from this research domain.

A closely related class of problems is \emph{parkour}, traditionally defined as the physical discipline of traversing complex routes in the most efficient manner possible. In robotics, the term has been adopted to broadly describe highly dynamic locomotion through environments composed of obstacles and gaps on the scale of the robot itself. To navigate these terrains, most recent approaches train \ac{RL} policies conditioned on explicit position commands~\citep{rudin2025parkour, hoeller2024parkour, vogel2025ladder, kim2025highspeed}. While our work uses a related goal-reaching reward structure, we do not condition the policy on goal positions. Instead, we encode the path itself in the terrain, so that the policy requires neither a goal location nor a command at deployment.

While most navigation and parkour tasks use geometric information readily obtained from range-based elevation mapping~\citep{miki2022elevationmapping}, estimating high-level traversability is fundamentally harder because it requires expert demonstrations or semantic information. In autonomous driving, one research field focuses on learning such top-down representations directly, lifting multi-view camera features into a common \ac{BEV} grid to predict dense semantic maps~\citep{philion2020lss, li2024fast, liao2025maptrv2}. This paradigm was recently extended to off-road traversability and elevation estimation~\citep{frey2024roadrunner}. In contrast, parameter-free geometric lifting has been shown to perform on par with learned strategies, with metric range sensing providing the dominant gains~\citep{harley2023simplebev}. A second line of research estimates traversability in a 2D space by leveraging human or robot demonstrations~\citep{jung2024vstrong, kim2024egotrav, aegidius2025stepp}. To further reduce the labeling burden, some approaches are capable of online adaptation~\citep{mattamala2025wvn, schreiber2025chungus}. However, their typical update rates of \SIrange{1.6}{8}{\hertz} render these methods unsuitable for high-speed racing tasks. Similar to~\cite{lucki2025vpe}, demonstrating high-rate segmentation, we decode the general-purpose features of a modern \ac{VFM} backbone into a semantically annotated image. We improve upon this baseline by pushing the rate of the complete semantic pipeline from \SI{50}{\hertz} to about \SI{90}{\hertz}, by adding a per-pixel confidence score, and by enabling adaptation to a new environment from 20 labeled images within minutes on the robot's own computer (subsection~\ref{ssec:semmap}).

A traversability estimator trained on a closed label set must also indicate when its predictions cannot be trusted. The maximum softmax probability~\citep{hendrycks2017baseline} often remains high on anomalous inputs~\citep{pinggera2016lostandfound, blum2021fishyscapes}, while Monte Carlo dropout requires multiple forward passes~\citep{kendall2017uncertainties}. Single-pass alternatives combine evidential learning with latent density estimates~\citep{ancha2024deep, cai2024evora} or refine anomaly scores with a promptable segmentation model~\citep{zhao2024segment}. Feature reconstruction offers a lighter signal, serving as segment-level confidence in~\cite{mattamala2025wvn} and as per-pixel uncertainty in~\cite{schreiber2025chungus}, the work closest to ours. We adopt this reconstruction branch but use cosine similarity to obtain a bounded score that directly weights map fusion, and exclude the background class from the reconstruction loss. Unlike softmax confidence, the resulting score flags unseen obstacles at a usable threshold (subsection~\ref{ssec:semmap}).

Navigating unstructured outdoor trails is frequently explored under the umbrella of ``autonomous hiking.'' Related methods typically do not focus on traversal time minimization and often rely on a hierarchical structure that bridges navigation with locomotion through a low-dimensional, human-engineered command interface~\citep{miki2023bridging, mattamala2025wvn, Reed2025, Giusti2016trail}. From an application perspective, this domain is well-represented by~\cite{miki2023bridging}, which introduces an architecture that uses a \ac{VLM} to convert RGB images every ten seconds into pairs of labels and risk values. An open-vocabulary segmentation network converts these pairs into a semantic map, which is then projected onto a 2.5D elevation map and subsequently consumed by two hierarchical networks. Compared to this and similar methods, we focus on traversal speed, achieving average velocities of \SIrange{3.0}{3.8}{\meter\per\second} and peak velocities above \SI{5}{\meter\per\second}. These traversal speeds are enabled by a unified \ac{MDP} formulation, procedurally generated trail terrains, and the high update rate of our mapping pipeline. Our comparison with hierarchical baselines (subsection~\ref{ssec:comparison_hierarchical}) indicates that a command interface, as used in these systems, limits reliability at high speed.

\section{Method}
We reinterpret both parkour and racing as instances of time-optimal traversal through structured or unstructured environments. Our approach does so by encoding the navigation task as a semantic layer within the terrain mesh (subsection~\ref{ss:terrain_gen}), providing this privileged information directly to the policy, and treating it as a soft boundary during learning (subsection~\ref{ssec:tean}). During deployment, we utilize a dedicated perception system to reconstruct this semantic layer in real-time (subsection~\ref{ss:semmap_method}).

The overall approach is presented in Fig.~\ref{fig:approach} and detailed in the following subsections.

\subsection{Terrain Generation} \label{ss:terrain_gen}
In prior work introducing \ac{UMV}~\citep{bokser2026umv}, symmetrical Isaac-lab terrains were utilized to learn a generic driving policy, where the robot starts in the center of a terrain patch and is tasked to follow a velocity command. In attempting to add an autonomy layer (e.g., by naively learning a high-level target), we encountered several problems with this formulation. First, randomly sampled twist and standing commands are difficult to align with the local terrain properties. Second, the twist commands influence the terrain curriculum and therefore the skill progression, leading to an entangled tuning problem between terrain topology and task rewards. Third, the agents learn more skills than required to drive down a trail, whereas the relevant skills are underrepresented for reaching peak racing performance. The method presented here resolves these pain points by introducing unidirectional terrains and embedding the directional and standing commands within the terrain. We train policies on two complementary classes of semantic terrain meshes: procedurally generated mountain bike trails designed to learn generic downhill skills, and hand-composed \acp{DT}, i.e., high-fidelity representations of real-world environments, for learning specialized parkour maneuvers. 

\subsubsection*{\acl{TL}}
\begin{figure*}[h!]
\centering
\includegraphics[width=\textwidth]{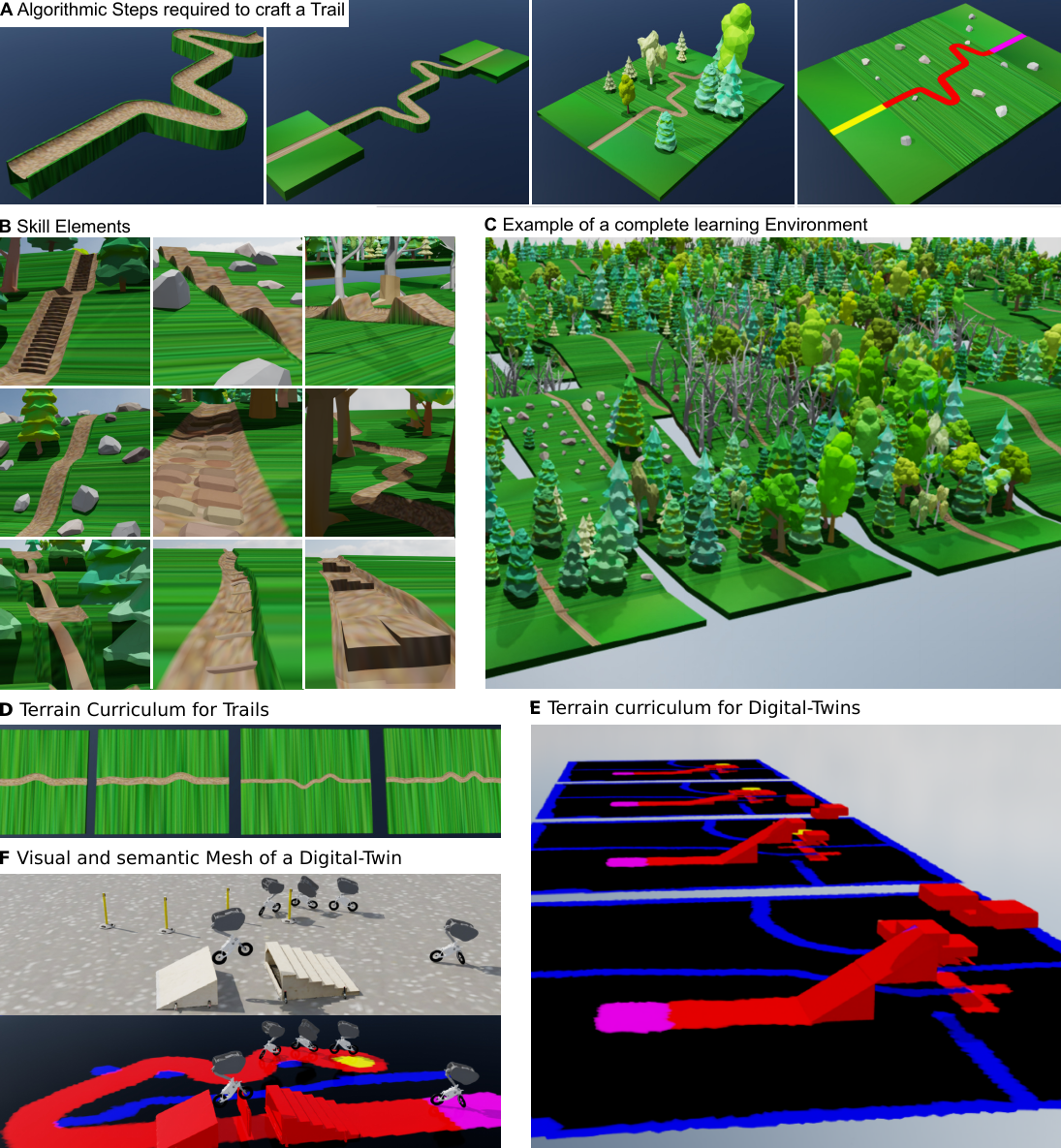}
\caption{\textbf{Mesh generation.} \textbf{(A)} The procedural generation of mountain bike meshes is divided into four stages. First, a cross-section polygon is swept along a 4D sweeping path ($x$, $y$, $z$, and yaw). Second, two geometric platforms are welded to the start and end of the trail. Third, the lateral vertices are extended to align with the terrain patch size. Forest objects can optionally be placed. Finally, semantic information is embedded into the vertex colors. \textbf{(B)} Each trail segment may contain one or more skill element(s) of a unique type, such as stairs, pump-track waves, jumps, drops, rocks, curves, skinnies, roots, and ramps. \textbf{(C)} Multiple terrain patches are instantiated across a uniform grid to construct the learning environment. \textbf{(D)} Example of the terrain curriculum for a curved segment. \textbf{(E)} Example of the height-based curriculum for parkour. \textbf{(F)} A high-resolution (\SI{17}{\mega\byte}) \emph{visual} mesh is constructed by composing pre-meshed objects. The corresponding \emph{semantic} mesh is generated by down-sampling the geometry to~\SI{5}{\mega\byte} and manually encoding high-level navigation signals into the vertex colors.
}
\label{fig:meshes}
\end{figure*}
The \ac{TL} is a modular framework for generating parameterizable mountain bike trails. Each terrain patch comprises four components: two platforms (the starting and ending zone), a connecting trail segment, and distinct skill elements. The resulting terrain is represented as a single mesh using the \texttt{trimesh} library~\citep{dawsonhaggerty2019trimesh}, which models geometry as triangles and stores visual information within the RGB channels. Further algorithmic details are illustrated in Fig.~\ref{fig:meshes}.

We maintain two parallel mesh representations: a semantically annotated mesh used for collision modeling and, optionally, a visually rich mesh for realistic rendering. Semantic information is explicitly stored in the vertex colors. The exact RGB values can be chosen arbitrarily, provided the mapping from RGB to semantic label remains bijective. In this work, we designate red $(1,0,0)$ for the primary trail, yellow $(1,1,0)$ for the starting zone, magenta $(1,0,1)$ for the final zone, and blue $(0,0,1)$ for illegal zones.

For the training environment, we generate 1000 randomized terrain patches, with each featuring two~\SI{10}{\meter} long platforms connected by a~\SI{36}{\meter} long trail, effectively yielding over~\SI{46}{\kilo\meter} of traversable tracks. The number of vertices per patch varies depending on the specific trail type, ranging from 2800 to 10\,000, summing to a total of approximately 4.5 million vertices for the combined training environment.

\subsubsection*{\acl{DT}:}
While the \ac{TL} offers an environment for learning generic mountain bike skills, it may not be well suited for learning specific parkour skills. We therefore introduce \acp{DT}, high-fidelity meshes replicating real-world parkour tracks. Visual meshes are constructed from a library of pre-meshed assets. Using a handheld capture rig, containing a stereo camera paired with a Livox Mid-360 LiDAR for \ac{LIO} state estimation, we record dense imagery of physical obstacles in our laboratory. We then reconstruct the geometry and texture of ramps, stairs, slopes, and poles into high-resolution meshes.

Next, we derive a semantic version of each \ac{DT} by manually embedding semantic classes directly into the RGB channels (``painting'' the desired path). We note that the specific geometry and the manual annotations substantially influence the robot's emergent behavior. To reduce simulation overhead, we minimize its polygon count using a quadratic edge collapse decimation algorithm~\citep{garland1997qem}. The process for designing the parkour mesh used for our control evaluation is detailed in Fig.~\ref{fig:meshes}F.

To further capture the realistic topography of natural outdoor environments, we also generate a large-scale \ac{DT} of our target mountain bike trail. We reuse the same capture rig to reconstruct the entire outdoor trail as a single, continuous mesh. Here, the semantic annotation is fully automated: we process the recorded camera streams through \ac{SemMap} and use the corresponding \ac{LIO} poses to project the predicted pixel classes directly onto the mesh vertices.

\subsubsection*{Terrain Curriculum}
Both the \ac{TL} and the \acp{DT} employ a terrain curriculum. An agent is promoted to a higher difficulty level if it successfully reaches the final zone within the episode, and demoted to the previous level otherwise. For the \ac{TL}, the parameterization of the skill elements scales linearly with the terrain level. Conversely, because \acp{DT} lack this parametric flexibility, we implement a height-scaling-based curriculum: At the lowest difficulty, all mesh vertices are projected onto a flat, two-dimensional plane. As the level progresses, the vertical dimensions incrementally scale until they reflect the true geometry. Two examples of the terrain curriculum are illustrated in Fig.~\ref{fig:meshes}D and~\ref{fig:meshes}E.


\subsubsection*{Terrain Randomization}
For the \ac{TL}, the terrain mesh is systematically randomized by overlaying the vertices with a sequence of geometric disturbances
\begin{equation}
    \vec v_i \gets \vec v_i + f_{\text{dist}, j}(\vec v_i, \text{difficulty}), \quad \forall j = 0,1,\dots
\end{equation}
where $f_{\text{dist}, j}$ can apply uniform noise, a constant slope, sinusoidal distortions, or other geometric perturbations. Additionally, decorative elements such as trees, roots, or stones can be placed adjacent to the trail. These perturbations break predictable structural patterns, preventing the policy from specializing to specific procedural trail outlines.

For the \acp{DT}, we apply both affine geometric transformations and semantic color randomizations. First, we construct an affine transformation matrix $A \in \mathbb{R}^{4 \times 4}$, initially set to the identity matrix. We then apply a random skew perturbation, $A[i, 0{:}3] \mathrel{+}= \mathcal{U}(-0.1, 0.1)$. Second, we embed the previously mentioned height curriculum directly into this transformation matrix, $A[2,2] = \text{difficulty}$. Third, to simulate boundary bleed effects, we randomly perturb the semantic labels of the mesh: If a vertex is located on a semantic boundary, its color is randomly flipped with a probability of $p_{\text{switch}} = 0.08$.

\subsection{\texorpdfstring{\ac{TENa}}{TENa} -- The Learning Environment} \label{ssec:tean}
Our racing and parkour policies are trained with on-policy \ac{RL} where all interaction data is gathered in simulation. The \ac{MDP} components remain identical for all tasks, unless noted differently. The core formulation is described in this subsection, while additional information can be found in the Appendix of sections \ref{supp:optimizer_settings} (optimizer settings), \ref{supp:sim} (simulation settings), \ref{supp:observations} (network architecture), \ref{supp:actions} (action space), \ref{supp:style} (style rewards), and \ref{supp:dr} (domain randomization).

\subsubsection*{Observations}
The policy network observes a vector of direct proprioceptive (i.e., state estimation independent) measurements. In addition, the policy receives a two-layer semantic height scan, as visualized in Fig.~\ref{fig:approach}B. The elevation layer informs the agent about the topological properties of the local surroundings. The semantic layer acts as a boolean indicator of whether a given scan point belongs to the traversable trail. Unlike goal-conditioned legged parkour frameworks, our policy does not observe the target location directly. Instead, it is only informed whether the agent has reached the final zone. This design choice is motivated by the fact that the time-optimal racing trajectory does not depend on the remaining distance to the goal, assuming long-horizon energy management is negligible. A list of observation space variables is provided in the Appendix~\ref{supp:observations}.

\subsubsection*{Commands}
Rather than relying on twist or pose commands, the agent autonomously discovers its motion while remaining softly constrained to the trail boundaries. This avoids sampling command targets that might conflict with the local terrain. The only command in use is a boolean standing command, which is activated when a single, gravity-aligned ray cast from the base position intersects a mesh-face labeled as a standing zone.

\subsubsection*{Rewards}
We distinguish between \emph{task} rewards, which define the core objective, and \emph{style} rewards, which are required to facilitate smooth sim-to-real transfer. The full formulation of the style rewards is detailed in the Appendix \ref{supp:style}. The primary racing task consists of three objectives: (i) forward exploration, (ii) remaining within the trail boundaries, and (iii) coming to a stable track-stand at the final zone. The total task reward is defined as the weighted sum:
\begin{equation}\label{eq:rewards}
r = w_1 r_1 + w_2 r_2 + w_3 r_3.
\end{equation}
High-speed racing emerges implicitly as a time-optimal behavior driven by the fixed episode length and the terminal standing reward. For all experiments and behaviors, we utilize the weights $w_1=1$, $w_2 = 0.2$, and $w_3 = 5$. 
\begin{itemize}
\item \textbf{Exploration} ($r_1$): To mitigate the sparsity of the terminal standing reward, we encourage forward progression during the active racing phase:
\begin{equation}\label{eq:r1}
    r_1 =
    \begin{cases}
    0, & c=1, \\
    \frac{v_{B,x}}{v_{\text{min}}}, & v_{B,x} \leq v_{\text{min}}, \\
    1, & v_{B,x} > v_{\text{min}}
    \end{cases}
\end{equation}
where $v_{B,x}$ is the base heading velocity expressed in the base frame, $v_{\text{min}}$ is the threshold velocity required to saturate the exploration reward, and the boolean indicator $c=1$ denotes that the agent has reached the final zone. We set $v_{\text{min}} = 2$ for all parkour and racing environments. The only exception is the high-speed Allmend trail simulation, where we used $v_{\text{min}} = 10$.\endnote{We note that for the high-speed simulation experiment, the achieved average speed is similar to the minimum velocity of the exploration reward, essentially rendering the exploration reward a task reward.}
\item \textbf{Survival} ($r_2$): To encourage the agent to remain within the valid track boundaries, we apply a dense survival reward:
\begin{equation}\label{eq:r2}
    r_2 =
    \begin{cases}
    1, & \text{alive} \land \text{on trail}, \\
    0, & \text{otherwise}.
    \end{cases}
\end{equation}
We found that framing boundary adherence as a soft dense reward rather than a hard termination criterion leads to a slightly higher success rate (see Appendix~\ref{supp:track_following}). To detect whether the agent is driving on the trail, we use a single, gravity-aligned ray to probe the terrain beneath it.
\item \textbf{Track-stand} ($r_3$): The agent is incentivized to come to a complete rest upon reaching the target zone using:
\begin{equation} \label{eq:track_stand}
    r_3 = e^{-\|\xi_B\|_1} \cdot c,
\end{equation}
where $\xi_B$ is the base twist expressed in the base frame.
\end{itemize}

Because the maximum standing reward per step ($w_3 + w_2$) exceeds the maximum reward collected while driving ($w_1 + w_2$), arriving earlier strictly increases the return within the fixed episode length, which induces time-optimal behavior without an explicit time objective or a path parameterization, as required by progress-based racing rewards. As discounting ($\gamma = 0.99$) attenuates this incentive far from the goal, the dense exploration reward $r_1$ provides the learning signal over long distances.

\subsubsection*{Terminations}
During training, episodes are terminated early under three specific conditions. First, an episode terminates if any joint violates its predefined position or velocity limits. Compared to \cite{bokser2026umv}, joint torque limits do not trigger termination; instead, they are enforced by clipping the output actuator signals before they are applied to the robot. Second, the agent is terminated if it navigates into designated illegal zones. These are user-defined semantic regions within the terrain mesh that act as hard boundaries, preventing the policy from exploiting shortcuts across the track. Third, an episode terminates if the robot's upper body makes physical contact with the semantic mesh environment. Actuator and joint position limits can be found in the preliminary work~\citep{bokser2026umv}.

\subsubsection*{Tuning}
We tuned all \ac{MDP} components to achieve speeds between \SIrange{3}{5}{\meter\per\second}, an interval that we found yields a good trade-off between real-world success rate and high-speed racing. We can achieve higher speeds by increasing the weight on the track-stand reward in Eq.~\ref{eq:track_stand} or decreasing the episode length. Once fixed, we do not change these components\endnote{The only exception is the simulated Allmend trail racing experiment, where we have increased $v_\text{min}$.} anymore, even when introducing a new task. Instead, we use the terrain mesh as a tool to shape the overall behavior of the robot. Two examples of shaping artifacts are shown in Fig.~\ref{fig:meshes}F. The beam placed within the gap forces the policy to learn an active jump rather than rolling through the gap, which would otherwise be the dominant discovered mode. The extended straight path connecting the turn with the entrance of the slope was designed as short as possible, but long enough to allow the robot to enter the slope in a fully upright position.

\subsubsection*{Training}
We trained and tuned all our policies on a single NVIDIA L4 GPU equipped with~\SI{24}{\giga\byte} of VRAM, utilizing 4096 parallelized environments. Notably, instantiating the entire curriculum of 1000 procedurally generated trail terrains within the Isaac ecosystem consumes approximately~\SI{4}{\giga\byte} of VRAM.

\subsection{\texorpdfstring{\ac{SemMap}}{SemMap} -- Real-World Perception} \label{ss:semmap_method}
\ac{SemMap} bridges the sim-to-real gap by producing the fixed-size, egocentric 2.5D semantic elevation map that the policy requires for real-world deployment. It comprises two pipelines (Fig.~\ref{fig:approach}C): First, a \emph{semantic pipeline} maps each RGB image to three per-pixel-aligned outputs: a segmentation image, a traversability image, and a confidence image. Second, a \emph{semantic elevation mapping} pipeline fuses the traversability and confidence images with the geometry built from point clouds and the robot pose. We describe the semantic pipeline and its label-efficient training, then the elevation mapping, and finally the system design that allows the whole pipeline to run in real-time and on-board.

\subsubsection*{Semantic Segmentation}
We cast the problem of semantic traversability estimation as supervised semantic segmentation followed by per-class scoring. As shown in Fig.~\ref{fig:approach}C, an RGB image passes through a frozen \ac{VFM} backbone, a feature upsampler, and a lightweight dual-head decoder that predicts per-pixel class probabilities together with a per-pixel confidence.

We extract features with a frozen DINOv3-s~\citep{simeoni2025dinov3} backbone. Such \ac{VFM} backbones provide dense, general-purpose features that transfer to downstream tasks through a lightweight decoder, here used for semantic segmentation. We resize the input RGB image to $320\times512$ pixels (height~$\times$~width), which the backbone tokenizes into a $20\times32$ patch grid. This resolution is large enough to retain the spatial details needed by the decoder, yet small enough to keep inference fast. The resolution also matches the camera's $10{:}16$ aspect ratio, avoiding the information loss of square cropping.

Patch tokenization reduces the features to a coarse $20\times32$ grid, blurring object boundaries. We recover fine details with a learned feature upsampler~\citep{couairon2025jafar} that conditions on the image's low-level cues to produce content-aware, sharper features. The inference cost of such upsamplers depends heavily on the guidance image size and the upsampling factor. We bound this cost in two ways: we use a downscaled guidance image rather than the full-resolution image, and we upsample only partially (by a factor of $4$ rather than the full $16\times$). The Appendix section~\ref{supp:upsampler} provides an ablation of how the upsampler affects inference speed and segmentation quality.

The decoder is kept deliberately lightweight because it is the only component trained per environment. Following~\cite{schreiber2025chungus}, it first projects the upsampled features from $384$ to $128$ channels and then splits into two heads. A segmentation head maps the $128$ channels to $N_c$ per-pixel class logits, where $N_c$ is the number of classes. A reconstruction head maps $128\to384$ channels back to the original backbone features and supplies the signal for confidence estimation.

Because the decoder is trained on a closed set of classes and a relatively small number of training examples, it will encounter unfamiliar terrain and objects at deployment. As a surrogate for epistemic uncertainty estimation, we obtain a per-pixel confidence from the cosine similarity between the original backbone features and those reconstructed by the reconstruction head. Where reconstruction succeeds, the content resembles the training distribution and confidence is high; where it fails, the pixel is likely \ac{OOD} and confidence is low. This yields an \ac{OOD} signal that captures epistemic uncertainty without any negative or \ac{OOD} examples during training, and we use it downstream to filter unreliable observations. We deliberately target epistemic rather than total uncertainty. Because the decoder is trained from a handful of labels, the dominant risk at deployment is unfamiliar content rather than label ambiguity. Aleatoric ambiguity, which concentrates at class boundaries~\citep{kendall2017uncertainties}, is instead absorbed downstream by design: the temperature-weighted softmax in Eq.~\eqref{eq:traversability} blends the traversability of ambiguous pixels rather than committing to a single class, and the confidence-weighted fusion averages out transient per-pixel noise as observations from successive frames accumulate in each map cell.

\subsubsection*{Traversability Estimation}
Segmentation classifies the surface type; traversability states whether the robot should traverse it. We keep these two concerns separate so that a single segmentation model can serve different robots and tasks. We assign each class $c$ a weight $w_c\in[0,1]$ and compute a per-pixel traversability $T_{i,j}$ as a temperature-weighted softmax over the class logits $P_{c,i,j}$,
\begin{equation}\label{eq:traversability}
    T_{i,j} = \sum_{c=1}^{N_c} w_c \, \frac{\exp(P_{c,i,j} / \tau)}{\sum_{c'=1}^{N_c} \exp(P_{c',i,j} / \tau)},
\end{equation}
where $\tau$ controls the sharpness of the class distribution. Because the weights $w_c$ act only on the decoder output, the same network can be reused across robots and scenarios by reassigning $w_c$, with no retraining.

\subsubsection*{Label-Efficient Training}
Only the decoder and the class weights are learned per environment; for adaptation to be practical, it must succeed from a handful of labels. We reduce the labeling burden on two fronts: selecting the most informative images to annotate, and generating their labels with minimal manual effort.

To choose which images to label from a large unlabeled pool, we extract the DINOv3-s class token of every image and build a FAISS~\citep{douze2025faiss} index over these embeddings. We cluster the embeddings into as many clusters as the label budget and label only the representative of each cluster, the image closest to its centroid, so the labeled set spans the semantic diversity of the data while staying bounded in size.

Public datasets are used with their existing annotations. For data collected in a new environment, we use a promptable open-vocabulary model, SAM3~\citep{carion2025sam3}. However, we do this only offline to generate labels, since running such a model online would be computationally costly for on-board inference. We prompt it with points or a text description of each target class to generate masks, which are then converted into per-pixel class labels for training. SAM3 can produce spurious masks when a prompted concept is absent, so a lightweight human filtering step removes incorrect masks before training.

We train the decoder with a cross-entropy loss on the segmentation head and a mean-squared-error reconstruction loss on the reconstruction head, keeping the backbone and upsampler frozen. A dedicated background class absorbs all unlabeled and irrelevant pixels, such as sky or space, which sharpens the decision boundaries of the task classes. We mask this background class in the reconstruction loss so that confidence is calibrated only on the semantics relevant to the task.

\subsubsection*{Map Fusion}
The policy is trained on a 2.5D semantic elevation map; hence, at deployment the perspective-view traversability must be lifted into the same egocentric world parametrization. We build on top of an existing GPU elevation mapping system~\citep{miki2022elevationmapping} and integrate the \ac{MEM} framework~\citep{erni2023mem} as a library, which lets the segmentation and mapping stages share GPU memory and avoids CPU/GPU round-trips. Geometry is obtained from point clouds together with the robot pose from onboard state estimation; following the \ac{MEM} update scheme, each grid-cell center is then projected into the traversability image to retrieve its traversability and per-pixel confidence, which are fused into the cell as described below. The egocentric map spans $15\times15$~m$^2$ at a cell size of~\SI{0.1}{\meter}.

Each cell is updated by confidence-weighted fusion: An observation's update weight grows with its predicted confidence and with its proximity to the camera. Each cell accumulates a capped evidence weight, so that well-supported cells resist being overwritten by sparse, low-confidence observations. Observations below a small confidence floor are skipped. This soft weighting, in place of a hard threshold, keeps the map stable during fast turns, when the camera briefly points away from the trail, or at grazing angles where segmentation degrades. The full update equations and parameters are stated in Appendix section~\ref{supp:fusion}. To keep the map egocentric without computationally costly shifting, we recenter it only once the robot moves past a distance threshold from the current center. Unlike a pre-mapped elevation map, online onboard mapping suffers from occlusions and a limited horizon. For the trail deployment, we therefore implemented an OpenCV-based inpainting filter, which inpaints both elevation and semantics based on grid cells that are observable in both depth and RGB. The joint inpainting allows us to flood-fill neighboring unobserved cells from adjacent observed cells by propagating geometric and semantic values (multimedia extension~\ref{mov:semmap}).

\subsection{Deployment Architecture and Real-Time Execution} \label{ssec:deployment_hardware}
The full autonomy stack runs on ROS2 Humble~\citep{macenski2022ros2} with Zenoh~\citep{corsaro2023zenoh} as the middleware, which in our deployments transported large messages such as images and point clouds more reliably than the CycloneDDS and FastDDS alternatives.

\subsubsection*{Control}
Our deployment stack leverages the ONNX runtime for efficient neural network inference, with policies exported via the \emph{exploy} framework~\citep{bellicoso2026exploy}. The trigger mechanism for the track-stand differs between deployment environments. During parkour, the standing command is inferred from the underlying ``virtual'' semantic map. For outdoor racing, however, the desired stopping locations would be arbitrary. To accommodate the standing feature, we bypass the semantic map and directly map a joystick button to the corresponding policy observation channel. Because this manual override logic is natively embedded into the exported ONNX graph via exploy, it requires no downstream modifications to the deployment code.

\subsubsection*{State Estimation}
Similar to~\cite{bokser2026umv}, for the indoor parkour experiments we use a \ac{MoCap} system to track the robot's 6D base pose within the world frame and sample the elevation map. In our case, this offline map is generated directly from the semantic mesh by projecting the geometry and semantics into an RGB and a depth image. These two images are then converted into a colored MuJoCo height-field, which is subsequently sampled by a ray-caster to produce a two-layered elevation map. 

For all other deployments, indoor and outdoor racing alike, state estimation runs onboard. We employ a custom \ac{LIO} pipeline, inspired by KISS-ICP~\citep{vizzo2023kissicp}. Our adapted method extends the baseline with voxel normals for robust scan alignment and provides a stable, robust state estimate at frame rates far exceeding those of input LiDAR data, which is important when traversing complex real-world environments at high speeds. The pipeline aligns incoming scans from a Livox Mid-360 LiDAR against an internal local geometric world representation to estimate a consistent~\SI{10}{\hertz} 6D pose. We fuse these poses with the raw IMU readings in a sliding-window factor graph to produce a high-frequency 6D pose estimate, which we use to motion-compensate both the primary Livox and the secondary Robosense E1R point clouds. Finally, the motion-compensated point clouds and pose estimates serve as input to \ac{SemMap}, while also getting merged into an internal \ac{LIO} local geometric world representation used for subsequent scan alignment.

\subsubsection*{Perception} \label{ssec:perception}
\ac{SemMap} receives three input streams over ROS2: RGB (\SIrange{30}{60}{\hertz}), gravity-aligned pose (\SIrange{30}{1000}{\hertz}), and point clouds from stereo or LiDAR (\SIrange{10}{30}{\hertz}). Perspective-view traversability alone requires only the RGB stream. To consume these high-rate streams in real time, we adopt the multi-process, shared-GPU-buffer architecture of the visual perception engine~\citep{lucki2025vpe}: a C++ front-end handles subscriptions and time synchronization, while the semantic and elevation-mapping pipelines run as separate GPU workers, keeping all data in GPU memory (Appendix \ref{supp:dataflow}). Finally, we export the networks from PyTorch to TensorRT via ONNX, which gives a 3- to 4-times speedup on the NVIDIA AGX Orin and raises the maximum throughput of the semantic pipeline, covering segmentation, traversability, and confidence, to about~\SI{90}{\hertz}. In deployment, we operate the camera at~\SI{30}{\hertz} and match the pipeline accordingly. The GPU-based (CuPy) elevation-mapping fusion node publishes the complete semantic elevation map at the same rate: the control policy reads the most recently published map at~\SI{50}{\hertz}, decoupling control from perception throughput. The~\SI{30}{\hertz} map-update rate is sufficient. At the peak speed of~\SI{5}{\meter\per\second}, successive map updates are separated by less than~\SI{0.17}{\meter}, under two grid cells, and geometry updates are bounded by the~\SI{10}{\hertz} LiDAR, whose range far exceeds the map extent. 

\subsubsection*{Hardware}
\begin{figure}[tbp]
\centering
\includegraphics[width=\columnwidth]{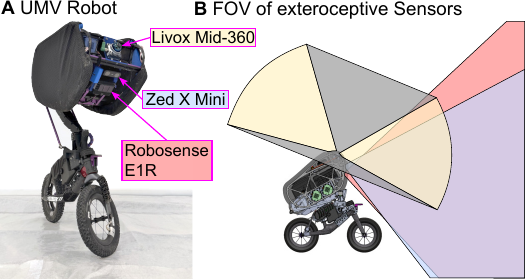}
\caption{\textbf{Hardware overview.} \textbf{(A)} The modified \ac{UMV} robot in a track-stand position showing the three different visual sensors. The robot has four actuated \acp{DOF}: the rear wheel, the fork, and the lower and upper torso joints, whereas the front wheel is unactuated. \textbf{(B)} The field-of-view of the cameras when the robot is in a resting position.}
\label{fig:hardware}
\end{figure}

We build on the \ac{UMV}~\citep{bokser2026umv}, a previously developed bicycle robot, and equip it with the exteroceptive sensors and compute required for onboard perception. Low-level control runs on the robot's existing onboard UP Xtreme i12, while perception and state estimation run on a separate NVIDIA AGX Orin 64GB added for this work. The sensor suite comprises a ZEDX Mini stereo camera that supplies RGB to \ac{SemMap}, a Livox Mid-360 LiDAR used primarily for \ac{LIO}, and a Robosense E1R LiDAR that provides the dense point cloud for elevation mapping. The full assembled robot is visualized in Fig.~\ref{fig:hardware}.

All sensors are mounted to a custom Kevlar roll cage, which has been ruggedized with waterjet-cut foam to withstand high-impact crashes during testing. The braided Kevlar core provides lightweight robustness, while the foam damping reduces shock loads on the sensors. The sensors themselves are mounted on a single 3D-printed CF-PLA support, enabling rapid positional iteration while maintaining alignment between sensors during testing. On the bike side, the native nylon covers shield the bike electronics from dirt and debris, while custom TPU bumpers on the fork and frame protect the carbon fiber from gouges. This robust design protects both the sensors and internal components from severe mechanical shock, minimizing downtime and enabling sustained, full-day testing during policy deployment.

\section{Experiments and Results} \label{sec:results}
The following results are structured to reflect our methodological progression, advancing from isolated simulation validation toward fully autonomous, high-speed real-world deployment. While all developed modules are conceptually platform-agnostic, we show the main results on the \ac{UMV} robot.

\subsection{Simulation} \label{ssec:simulation}

We first verified that \ac{TENa} generalizes across a diverse set of \acp{DT} by showing different emerging behaviors for parkour and racing tasks. To this end, we trained specialized policies in each environment and evaluated their performance, measured by success rate. During training, we enabled all domain randomizations required for real-world deployment, whereas for evaluation, these randomizations were disabled to establish nominal performance. In the appendix~\ref{supp:anymal}, we further demonstrated cross-embodiment generalization in simulation by applying our identical learning framework to the ANYmal-D quadruped for both trail and parkour.

\subsubsection*{Parkour}
\begin{figure*}[h!]
\centering
\includegraphics[width=\textwidth]{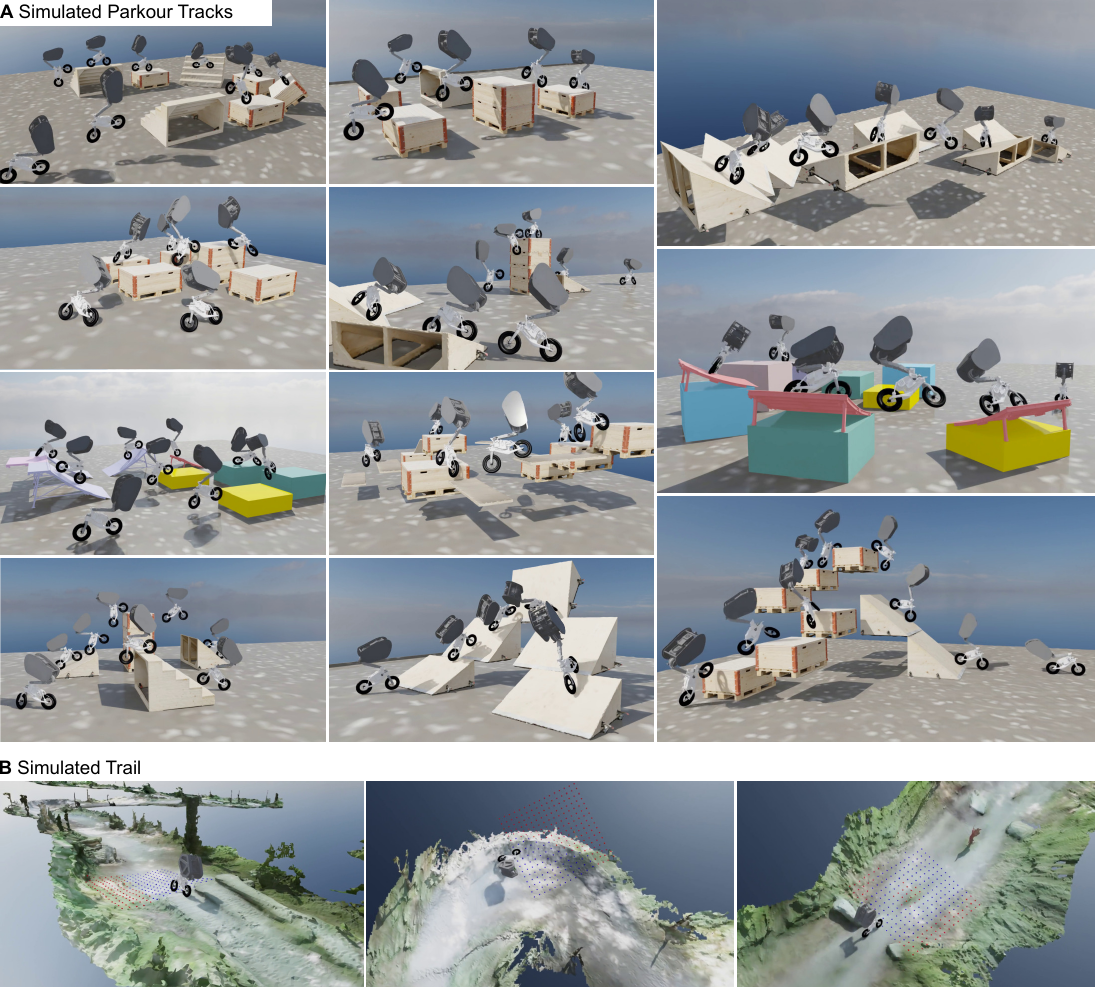}
\caption{\textbf{Parkour and trail racing learned using \ac{DT} meshes.} All tracks share a similar core concept: Semantic information is embedded into the mesh, creating a unique trail connecting two virtual platforms, the starting and the final zone. This trail is made observable to the policy and used to shape the gradient landscape. \textbf{(A)} The time-lapse images show individual parkour tracks together with the emerging locomotion behavior. \textbf{(B)} High-speed racing on a reconstructed real-world mountain bike trail, reaching average speeds of \SI{9.81}{\meter\per\second}. The colored dots illustrate the annotated height scan, with blue labeling the trail and red labeling non-trail.
}
\label{fig:sim}
\end{figure*}
We designed 17 parkour tracks featuring stairs, wall-rides, gaps, drop-ins, ramps, and boxes. These tracks serve two purposes: First, they let us design and tune a general \ac{MDP} formulation so different tasks can be solved through a shared interface. Second, once implemented in the real world (see subsection~\ref{ssec:parkour}), these tasks allow us to identify the control sim-to-real gap in terms of measurable success rate. 

We found that our \ac{MDP} formulation successfully discovers completing behaviors for all environments, i.e., behaviors that solve the last difficulty level with non-zero success rate. We used only episode length as a tuning parameter (see Appendix~\ref {supp:optimizer_settings}). Across 1000 simulated runs with randomized initial conditions, we report success rates between~\SI{95}{\percent} and~\SI{100}{\percent}, depending on the parkour type. A run is deemed successful only if the robot navigates the entire course and smoothly transitions into a stable track-stand on the final zone. A time-lapse of 11 selected parkour environments is presented in Fig.~\ref{fig:sim}A, whereas in multimedia extension~\ref{mov:sim_parkour} we show the complete set of 17 parkour tracks. 

Compared to previous work~\citep{bokser2026umv}, we extend the skill set of \ac{UMV} from track-stand, active jumps, hops, drive, and flips to further include cornering, wall-rides, gaps, drop-ins, hip-jumps (jump and yaw), stair landing, and slaloms. We further increase the active jump height from \SI{1}{\meter} to \SI{1.75}{\meter}; however, we note that this result may not transfer zero-shot to the real robot due to operation close to both joint and torque limits. 

\subsubsection*{Trail}
%
%
Our target deployment mountain bike trail, the \emph{Allmend trail}, measures~\SI{969}{\meter} in length with a total elevation drop of~\SI{59}{\meter} and a maximum steepness of~\SI{-18}{\degree}. At the time of writing, the all-time Strava leaderboard for this combined segment reported a record of~\SI{105}{\second} (averaging~\SI{9.5}{\meter\per\second}), with an all-time median of~\SI{5.3}{\meter\per\second} across 2861 human riders.\endnote{Strava: https://www.strava.com/segments/29736346, accessed 03 June 2026}. Due to a pedestrian hiking path, the mountain bike route is artificially divided into two parts. We reconstructed a high-fidelity \ac{DT} of the upper segment, which spans a distance of~\SI{419.1}{\meter} with an elevation drop of~\SI{21}{\meter} and a maximum steepness of~\SI{-5.4}{\degree}. The \ac{DT} captures all skill elements, including a wall-ride, two drop zones, a seesaw, a rock bed, two skinnies, a slalom section, and several root networks (Fig.~\ref{fig:trail_racing}).

We trained a specialized policy on this \ac{DT} and evaluated this policy in simulation across 1000 trials with randomized initial conditions. Due to the nature of the training setup, we will refer to the resulting policy as ``specialized racing policy''. Performance was quantified by the time required to reach the end of the trail and the overall success rate. Only those agents contribute to the total racing time that survive the episode. The agents completed the trail with an average time of~\SI{42.7}{\second} and an~\SI{89}{\percent} success rate. Assuming the traveled distance matches the official segment length provided by Trailforks\endnote{https://www.trailforks.com/trails/allmendtrail/}, the agent maintained an average racing speed of approximately~\SI{9.8}{\meter\per\second}. While our results suggest that \ac{TENa} can, in simulation, reach speeds comparable to the fastest human riders, this comparison is indicative only: the Strava times cover the entire trail, whereas our \ac{DT} covers the flatter upper section. Any behavior robustification typically counteracts peak performance, and thus, we do not expect this policy to transfer to the real world. Some snapshots of the simulated run are provided in Fig.~\ref{fig:sim}B, a summary of the results is shown in Table~\ref{tab:allmend}, and multimedia extension~\ref{mov:sim_allmend} visualizes the entire run in a top-down view.


If episode durations are relatively large, as in the case of the Allmend trail (Tab.~\ref{tab:ppo}), the standing reward becomes more sparse in time and hence, more difficult to discover. We found that increasing the minimum velocity instead provided a better handle for tuning the base velocity. However, this only works for tracks that can be entirely traversed at high speeds and do not require large velocity leaps, as, e.g., the parkour tracks. Consequently, to achieve the large velocities in this particular experiment, we increased the minimum velocity in the reward function~\ref{eq:r1} from \SI{2}{\meter\per\second} to \SI{10}{\meter\per\second}.

Without this change, the traversal time increases to~\SI{75.2}{\second} (corresponding to \SI{5.57}{\meter\per\second} average speed) and the success rate reaches \SI{100}{\percent}. 

\begin{table}
\caption{\textbf{Performance comparison between human riders (real-world data) and \ac{TENa} (simulated data) on the Allmend trail.} Human velocity metrics were retrieved from Strava. To further demonstrate generalization across diverse systems, we also include the traversal metrics of ANYmal. We note that the simulation runs are averaged over the upper part of the Allmend trail, whereas Strava provides measurements averaged over the entire trail.}
\centering
\begin{tabular}{lccc}
\hline
Agent &  Time & Velocity & Success Rate \\
\hline
Human (Med.) &    &\SI{5.3}{\meter\per\second} & unknown \\ 
Human (Rec.) &    &\SI{9.5}{\meter\per\second} & unknown \\ \hline
\ac{UMV} & \SI{42.7}{\second}  & \textbf{\SI{9.81}{\meter\per\second}} & \SI{89}{\percent} \\
ANYmal  & \SI{89.5}{\second}   & \SI{4.68}{\meter\per\second} & \SI{81}{\percent } \\ \hline
\end{tabular}
\label{tab:allmend}
\end{table}

\subsection{\texorpdfstring{\ac{MoCap}}{MoCap} Assisted Parkour} \label{ssec:parkour}
In the second phase of validation, we isolated and bridged the control sim-to-real gap for the \ac{UMV} by deploying it in a perception-independent environment. This isolated setup allows us to quantify the gap by comparing the success rates achieved in simulation against those on the physical track. To this end, we trained a policy on the parkour track displayed in Fig.~\ref{fig:meshes}F, one of the 17 parkour meshes. It features a four-pole slalom, a slope for lift-off, and a staircase for touch-down. The maximum height of both the slope and the stairs is~\SI{0.555}{\meter}. The slope has an inclination of~\SI{24.8}{\degree}, while the stairs have a step width of~\SI{0.27}{\meter} and a step height of~\SI{0.09}{\meter}. A gap of~\SI{1.2}{\meter} separates the slope and the stairs. These elements are specifically designed to abstract the core challenges of our target mountain bike trail (e.g., tight curves, jumps, and uneven ground) into a controlled indoor environment. 

\begin{figure*}[h!]
\centering
\includegraphics[width=\textwidth]{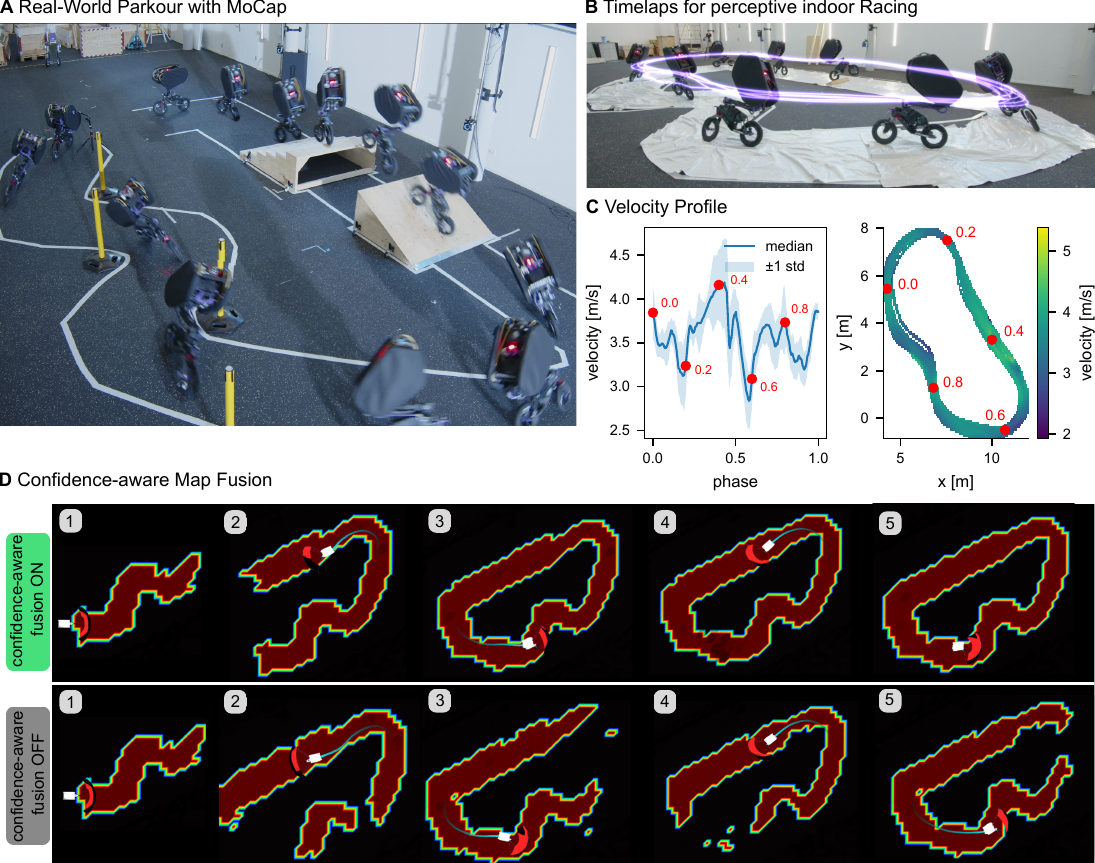}
\caption{\textbf{Indoor parkour and indoor racing.} \textbf{(A)} \emph{\ac{MoCap}-assisted indoor parkour.} The parkour track features a slalom segment, a ramp for lift-off, and a staircase for landing. \textbf{(B)} \emph{Autonomous indoor racing using onboard perception.} The \ac{UMV} navigates a high-contrast circuit utilizing only full onboard compute and perception, completing 130 consecutive laps over a duration of~\SI{891}{\second}. \textbf{(C)} \emph{Telemetry of the indoor racing experiment.} The left plot illustrates the median base velocity with $\pm 1$ standard deviation. The right plot displays a 2D positional trace of all laps overlaid with a velocity heatmap. Red markers denote five phase positions ($0$: starting location, and $1$: full lap). \textbf{(D)} \emph{Confidence-weighted map fusion during indoor racing.} Top: With confidence-weighted fusion. Bottom: Low-confidence observations from high-speed turns progressively corrupt the map.
}
\label{fig:indoor}
\end{figure*}

To break the dependency on perception, we used a \ac{MoCap} system to localize the robot within a pre-mapped \ac{DT}. For each test, the robot is initialized at the starting location in an upright position. Once the controller was engaged, the robot autonomously raced through the track and navigated to the dedicated stopping location. A representative successful run is visualized in Fig.~\ref{fig:indoor}A, and four successful runs are shown in multimedia extension~\ref{mov:mocap_parkour}. The robot entered the slope at a peak base velocity of~\SI{5}{\meter\per\second}, lifted off at~\SI{3.3}{\meter\per\second}, remained airborne for~\SI{0.4}{\second}, re-accelerated to~\SI{5}{\meter\per\second} on the stairs, and decelerated within~\SI{0.5}{\second} to a stable track-stand, completing the sequence in~\SI{8}{\second}.

Out of 10 consecutive trials on our final test day, the robot completed the run with a stable track-stand 8 times. We recorded two falls, both occurring immediately after the landing phase on the stairs. We identified the landing as the most fragile phase of the track, as success mostly depends on beneficial initial conditions set at lift-off. Due to a lack of a roll joint in the upper body, the robot cannot recover from large roll angles after landing. For comparison, in the equivalent simulated parkour track, we achieved a~\SI{99}{\percent} success rate, where near-optimal lift-off conditions are usually achieved. The difference in success rate is what we consider to be the sim-to-real gap.

Throughout the development, we have observed three other failure modes not present in simulation: (i) If the robot accelerates too fast into the turn before lift-off, it tends to slip off the ramp and fall into the gap. (ii) If the robot drives too fast over the slope, it may land on the flat ground, completely clearing the staircase. The resulting impact forces are typically too large for the upper body to absorb, leading to joint position or torque violations. (iii) If the robot drives too slowly over the ramp, the rear wheel will hit the front edge of the stairs before landing. While this strategy works in simulation, it leads to violent hardware destruction in the real world (e.g., exploded tires, broken rims, or even broken carbon frames). Those issues were mitigated through careful domain randomization (Appendix section~\ref{supp:dr}) and track tuning over the course of several months, aimed at rendering the policy sufficiently conservative to not excite these failure modes, but aggressive enough to complete the track.   

``Table jumps'' were previously demonstrated in \cite{bokser2026umv}, showing \ac{UMV} jumping on top of a table and then driving down from it. Both setups use \ac{MoCap} and human-designed tracks; ours replaces waypoint following with the unified formulation and adds a slalom, a gap and a staircase landing. We note that the rewards and tuning parameters proposed for the ``table jumps'' led to slightly conservative behaviors failing our parkour track, but still provided a meaningful starting point for our tuning process (Appendix~\ref{supp:dr}).

\subsection{Evaluation of Perception} \label{ssec:semmap}
We evaluate the \ac{SemMap} pipeline in isolation, quantifying its robustness, confidence signal, label efficiency, inference speed, and generalization across environments and tasks.

\subsubsection*{Qualitative Evaluation}
\begin{figure*}[h!]
\centering
\includegraphics[width=\textwidth]{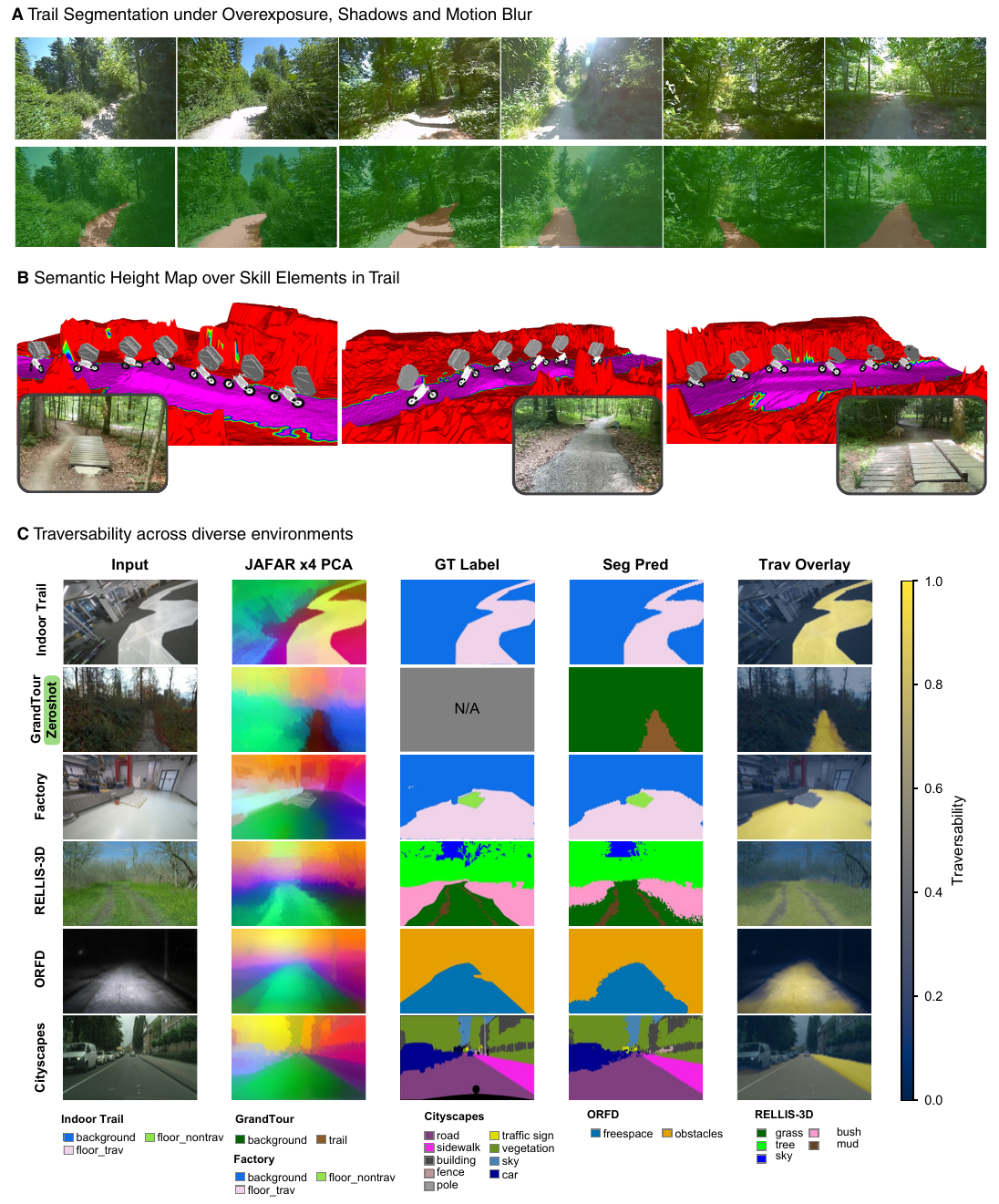}
\caption{\textbf{Real-world semantic perception with SemMap.} \textbf{(A)} Input frames (top) and segmentation overlays (bottom) under glare, cast shadows, and motion blur. \textbf{(B)} Semantic elevation map over a seesaw, pump-track waves, and a wooden tabletop. Insets: reference photos, not onboard. \textbf{(C)} Results over six datasets. The frozen DINOv3-s backbone and upsampler are shared; only the lightweight decoder is trained per dataset. GrandTour is evaluated zero-shot, hence no ground-truth labels (N/A). Rows, columns, and legends are labeled in-figure.}
\label{fig:perception_qualitative}
\end{figure*}

We assessed \ac{SemMap} qualitatively across diverse visual conditions and environments; Figure~\ref{fig:perception_qualitative} shows this characterization, and multimedia extension~\ref{mov:semmap} presents the pipeline, the cross-domain and zero-shot evaluations, the inpainting and fusion mechanisms, and the new-environment adaptation workflow in video form. On natural trails, the input imagery is subject to harsh sunlight, glare, cast shadows, and motion blur, yet the trail remains segmented in each of these instances (Fig.~\ref{fig:perception_qualitative}A). The resulting per-pixel traversability projects into a coherent semantic elevation map over the trail's skill elements (Fig.~\ref{fig:perception_qualitative}B). We further evaluated across six datasets spanning natural trails, an indoor racing track, an industrial floor, and off-road and urban scenes (Fig.~\ref{fig:perception_qualitative}C). Across all six domains, only the lightweight decoder is trained per dataset while the DINOv3-s backbone and upsampler stay frozen and shared, showing that one pipeline recipe transfers across natural, indoor, industrial, off-road, and urban scenes. In each case, a \ac{PCA} of the JAFAR-upsampled~\citep{couairon2025jafar} features already separates the scene into semantically meaningful regions with sharp object boundaries, and the decoder maps these features to task classes, from which a weighted softmax layer produces the per-pixel traversability.

\subsubsection*{Confidence and \acs{OOD} Detection}
\ac{SemMap}'s per-pixel confidence enables confidence-weighted map fusion, down-weighting unreliable observations before they enter the map, which is essential at racing speed. We demonstrated this on the robot during indoor racing (Appendix section~\ref{supp:ood}, Fig.~\ref{fig:indoor}D): with fusion enabled, the semantic elevation map stays consistent in each lap. With it disabled, the low-confidence predictions produced when the camera swings away from the trail in fast turns progressively corrupt the accumulated map. We do not target state-of-the-art \ac{OOD} detection, but a signal that is nearly free at inference, needs no \ac{OOD} examples, and is retrained with the decoder; we therefore compare against the softmax confidence of standard segmentation networks. Because no standardized \ac{OOD} benchmark exists for off-road trails, we further quantified the underlying confidence signal on the widely used Lost and Found benchmark~\citep{pinggera2016lostandfound}, which contains real lost-cargo obstacles on urban roads. Trained on Cityscapes~\citep{cordts2016cityscapes} and evaluated at three classes on the $1{,}111$ test images that contain obstacles, the \ac{SemMap} decoder predicts~\SI{35}{\percent} of obstacle pixels as traversable before any confidence filtering, compared to~\SI{60.2}{\percent} for DeepLabV3+~\citep{chen2018deeplabv3plus} and~\SI{69.4}{\percent} for SegFormer-B0~\citep{xie2021segformer}, i.e., $1.7\times$ and $2.0\times$ fewer (per-image paired Wilcoxon $P<0.001$, rank-biserial $r=0.76$--$0.85$). Its confidence further reaches a \SI{95}{\percent} true-positive rate at a threshold of 0.72, whereas both softmax baselines require 1.0, leaving no tunable range. This confidence-weighted fusion reduces the perception errors the policy must tolerate at deployment.

\subsubsection*{Label Efficiency}
Adapting \ac{SemMap} to a new environment is computationally cheap, as the decoder is the only component trained per environment. To quantify this label efficiency, we evaluated on the off-road RELLIS-3D benchmark~\citep{jiang2021rellis3d}. At a budget of 20 labeled images, our frozen-backbone decoder ($\sim$100\,000 trainable parameters) reaches $33.9 \pm 1.4\%$ \ac{mIoU} (mean $\pm$ SD, $n=3$ seeds), whereas the fully fine-tuned baselines reach $23.9 \pm 0.9\%$ (DeepLabV3+) and $24.1 \pm 3.5\%$ (SegFormer-B0); with 100 labeled images, only ~\SI{3}{\percent} of the training set, the decoder already matches its own full-dataset performance. These absolute values are low because the benchmark's 20 classes are imbalanced; pixel accuracy exceeds~\SI{88}{\percent} even at the lowest budgets. On our own outdoor trail dataset, 20 labeled images reach $82.6\%$ \ac{mIoU} on the held-out test split, while the full training set of 28\,763 images reaches \SI{90.0}{\percent}, indicating that the additional data buys robustness margin rather than base functionality. Because the decoder converges in a few epochs on frozen features, adaptation can run on the robot's own computational resources, an NVIDIA AGX Orin: starting from a 5-minute recording of an unseen environment, we produced a deployed model in under 15 minutes of combined human and compute time, comprising 2 minutes of automated data curation, under 5 minutes of SAM3-assisted labeling~\citep{carion2025sam3}, and 4 minutes of on-device training and TensorRT export. The full study is shown in the Appendix (section~\ref{supp:label_efficiency}).

Three factors explain why the frozen-backbone decoder outperforms the fine-tuned baselines when few labeled images are available. First, the decoder trains approximately 100\,000 parameters on top of the frozen backbone and upsampler, whereas the fine-tuned baselines adapt 3.7M and 42.0M parameters from the same 20 images; the gap narrows at larger budgets (Fig.~\ref{fig:label-efficiency}), indicating a difference in sample efficiency rather than capacity. Second, the gain depends on the pretrained representation rather than on freezing alone, as the frozen ResNet-50 is the weakest configuration. A \ac{PCA} of the upsampled DINOv3 features already separates the scene into class-consistent regions (Fig.~\ref{fig:perception_qualitative}C), such that the decoder acts as a near-linear readout, and the same structure is what renders the reconstruction-based confidence informative. Third, additional labels improve robustness rather than base performance: on the outdoor trail dataset, increasing the training set from 20 to 28\,763 images raises the \ac{mIoU} from \SI{82.6}{\percent} to \SI{90.0}{\percent} with diminishing returns (Table~\ref{tab:trail-label-efficiency}), suggesting that semantic coverage through clustered image selection matters more than label volume.

\subsubsection*{Inference Speed}
The pipeline must keep pace with a robot moving above~4~m\,s$^{-1}$. Our recommended JAFAR$\times4$ configuration runs model inference, including the confidence head, at~\SI{99}{\fps} (mean over $n=100$ timed iterations, latency SD $\le$~\SI{0.33}{\milli\second}), while improving \ac{mIoU} on the ORFD test set~\citep{min2022orfd} by 4.1 percentage points over bilinear upsampling. Higher upsampling factors add under 0.2 percentage points while reducing inference speed to roughly a third. Including input preprocessing, the traversability computation, and upsampling of all outputs to input resolution, the complete semantic pipeline runs at about~\SI{90}{\hertz} (\SI{11.1}{\milli\second} per frame), or at about~\SI{163}{\hertz} with bilinear upsampling (Appendix section~\ref{supp:upsampler}).

\subsubsection*{Generalization}
To probe generalization, we apply the decoder, trained only on our own outdoor trail data, zero-shot to the TRIM-1 sequence of the Grand Tour dataset~\citep{frey2026grandtour}, a forest gravel trail recorded under cloudy late-autumn conditions by an ANYbotics ANYmal-D quadruped, differing from our recordings in sensor, platform, viewpoint, season, and location. Since no labels exist for this location, the assessment remains qualitative: the decoder still produces coherent traversability, indicating that a single decoder can serve a broad class of trails rather than a single site. We further adapted the decoder to three unseen Grand Tour sequences, an alpine snow field, an indoor building, and an ice cave, each from 20 clustered, SAM3-labeled images. Trained on the AGX Orin in \SIrange{1.5}{3}{\minute}, the decoders reach \SIrange{77}{91}{\percent} \ac{mIoU} on held-out frames (Appendix Section~\ref{supp:label_efficiency}), confirming that the workflow yields a functional decoder on new sites.

\ac{SemMap} is robot-agnostic by design, observing the world only through the camera; its training data was collected primarily with a handheld capture device and, in some cases, a smartphone, independent of the deployment robot. Adapting only the decoder, we additionally deploy it RGB-only on a Unitree G1 humanoid for taped no-go-zone avoidance in a factory, where traversability is defined by the tape boundary rather than the surface and is thus invisible to geometry (Appendix Section~\ref{supp:g1}).

\subsubsection*{Limitations}
These deployments also delineate the limits of our design choice. The zero-shot Grand Tour transfer shows that a decoder generalizes across trails of a similar appearance; however, trained on a small, closed dataset, it does not generalize universally, and sustained appearance shifts, even within the same environment, are handled by adding samples from the shifted distribution, which the label-efficient training makes cheap. In practice, we extended the training set once: as the vegetation changed with the season, we recorded additional footage and retrained the decoder through the same rapid adaptation workflow (Appendix section~\ref{supp:deployment_days}). We also observed transient segmentation errors, such as the trail briefly appearing to fork into two paths for one or two frames; none of these required intervention.

\subsection{Indoor Racing} \label{ssec:trail_indoor}
Our next objective was to close the control-and-perception loop. We approached the trail racing challenge by training a racing policy across~\SI{46}{\kilo\meter} of synthetic trails generated by the \ac{TL}. As this policy was not ``specialized'' to a particular track as done in the previous experiments, we refer to it as ``generic racing policy''.

For physical deployment, we constructed a narrow racing track using high-contrast white fleece, featuring two sharp turns and a slalom segment, essentially outlining the previous parkour track without the slope and the stairs. Additionally, we introduced a~\SI{0.1}{\meter} tall wooden incline just before the first turn (Fig.~\ref{fig:indoor}B). The \ac{MoCap}-state estimator and ground-truth map are replaced with our LIO solution and \ac{SemMap}. The robot was initialized inside the track boundaries, and the racing controller was engaged from a track-stand position. In total, the \ac{UMV} completed 130 continuous laps, maintaining an average speed of~\SI{3.54}{\meter\per\second} and reaching a peak velocity of up to~\SI{5}{\meter\per\second} at the end of the straightaway. The average lap time measured was \SI{6.85}{\second}. Further details are provided in Fig.~\ref{fig:indoor}C, whereas multimedia extension~\ref{mov:perceptive_indoor_racing} documents all laps.

This experiment allowed us to further tune additional \ac{MDP} components related to the perception interface (e.g., the noise model in Appendix~\ref{supp:observations} or the terrain distribution), and to quantify the sim-to-real gap. In simulation, with all domain randomization terms disabled, we achieved a lap time of \SI{6.1}{\second}, as opposed to \SI{6.85}{\second} in the real world. We attribute the difference in speed to unmodeled or inaccurate actuator dynamics and rolling resistance. 

Due to the controlled indoor environment, the missing occlusions, and the flat ground simplification, we expect the results to constitute an upper bound of performance for real-world mountain bike trail deployment. We note that this generic racing policy was not capable of traversing the simulated Allmend trail due to the appearance of narrow slalom poles, which do not show up in the height scan, but lead to termination upon impact. As during the time between mapping and deployment, the ``slalom'' segment was restructured to a ``wave'' segment (consisting of seven pump-track waves), we did not attempt to solve this issue, and instead, directly deployed the policy on the Allmend trail.

\subsection{Trail Racing} \label{ssec:trail_racing}
\begin{figure*}[h!]
\centering
\includegraphics[width=\textwidth]{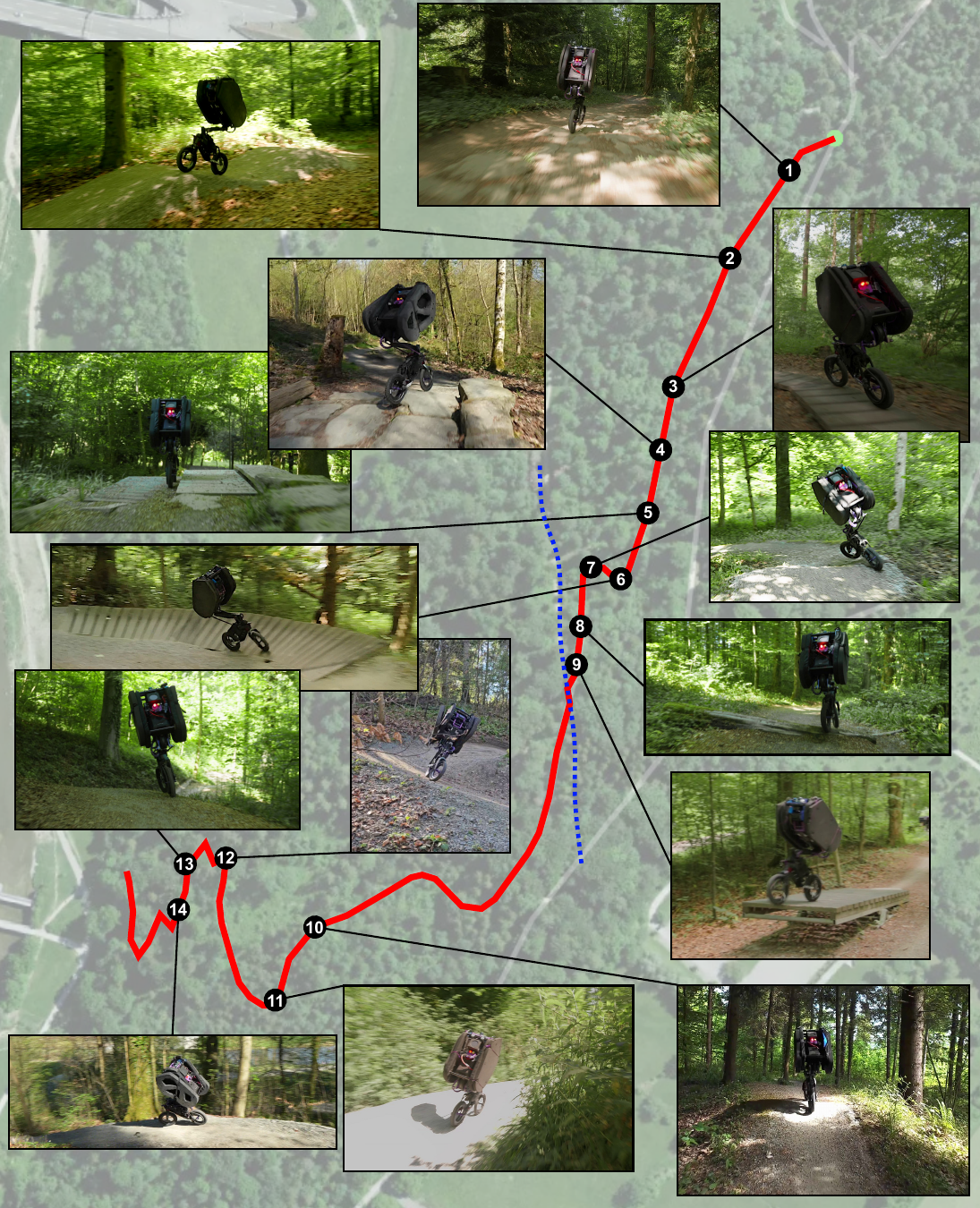}
\caption{\textbf{Deployment on a mountain bike trail.} The Allmend trail, a kilometer-long technical mountain bike route located in Zurich, is outlined in red; square/circle markers denote the start/end location. The dotted blue line marks the pedestrian path that cuts the trail into two segments. The \ac{UMV} navigates diverse skill elements at the numbered checkpoints: (1) rock bed, (2) sequence of seven waves, (3) two skinnies, (4) rocky tabletop drop, (5) wooden tabletop drop, (6) wall-ride, (7) tight turn, (8) natural transversal log, (9) locked seesaw forming a ramp, (10) natural hill, (11) extended and overgrown curve, (12) S-shaped descending curve, (13) steep descent, and (14) cliff-like trail segment with a sharp drop-off on one side.
}
\label{fig:trail_racing}
\end{figure*}
We deployed our fully integrated pipeline on the Allmend trail in Zurich using the generic racing policy synthesized previously. As outlined in Fig.~\ref{fig:trail_racing}, the upper section features wide areas populated with human-made skill elements and natural obstacles. In contrast, the lower section transitions into a flow-trail, characterized by several waves, steep descents, and narrow segments that are heavily overgrown with dense vegetation. In the following, we summarize the results from three independent successful deployment days. The reported mean velocities are computed from the norm of the state estimator's base velocity, and traveled distances are estimated as the integral of this value.

Our proposed racing framework traversed the Allmend trail across six continuous runs. These runs are successful runs that cover the entire trail, selected from a set of 79 runs, with a median traveled distance of about \SI{74}{\meter} (excluding the second deployment day; see Appendix section~\ref{supp:deployment_days} for a complete analysis). The longest continuous segment of this list measured~\SI{220}{\meter} and was traversed with an average velocity of~\SI{3.8}{\meter\per\second}, reaching a peak velocity of~\SI{5.2}{\meter\per\second}. A complete breakdown of the individual runs can be found in Table~\ref{tab:allmend_real}. Accordingly, the \ac{UMV} completed~\SI{946}{\meter} of real-world mountain bike trail across six runs, sustaining autonomous racing speeds above~\SI{3}{\meter\per\second} consistently. The longest autonomous run measured \SI{76}{\second} (Appendix section~\ref{supp:deployment_days}), which is about 12 times shorter than the autonomous run achieved indoors. The difference reflects the complexity of an outdoor deployment in wild forests. We provide a summary of all failure cases in the following subsection~\ref{ssec:fail}.   

Based on video footage in~\cite{bokser2026umv}, we estimate that the human-operated \ac{UMV} averaged \SI{1.36}{\meter\per\second} on the upper section of the trail.\endnote{We divided the length of the trail \SI{419}{\meter} by the measured time that passed between the point of entrance and the point of exit. We started the counter as soon as the robot passed the starting location used in our experiment (time \SI{18}{\second}) and stopped the timer when it reached the point where we stopped our experiment (time \SI{115}{\second}). We also compensated for the speedup factor of four applied between time \SIrange{23}{93}{\second}, leading to a total estimate of $(23-18)+ 4\cdot (93-23) + (115-93)=5+280+22=\SI{307}{\second}$. The average velocity comes out to \SI{1.36}{\meter\per\second}.} Averaged over this same segment, our autonomous policy was 2.3 times faster.\endnote{For comparison, we only consider the average velocity achieved on the upper trail.} We attribute the performance difference to two factors: the removal of the intermediate command interface and a training distribution tailored to the task. A navigation policy learned on the same trail terrains reduced, but did not close, this gap (subsection~\ref{ssec:comparison_hierarchical}), indicating that the command interface itself accounts for a substantial part of the difference.

\cite{bokser2026umv} also demonstrated velocity tracking of up to \SI{8}{\meter\per\second} on flat indoor ground under teleoperation, which marks the hardware limit of the \ac{UMV}. On the trail, our policy operated well below this limit, at average speeds between \SI{3.0}{\meter\per\second} and \SI{3.8}{\meter\per\second} per run. Here, speed is bounded not by the hardware but by the terrain, the perception horizon, and the trade-off between speed and robustness discussed in subsection~\ref{ssub:sim2real_gap}.

\begin{table}
\caption{\textbf{Performance on the real-world Allmend trail.} The table shows three runs on the upper and three runs on the lower part. The checkpoints refer to the skill elements shown in Fig.~\ref{fig:trail_racing}. Velocities are computed as the mean over the run. The total average velocities are computed as the total distance traveled divided by total time traveled.}
\centering
\begin{tabular}{llccc}
\hline
part & checkpoints & velocity & distance & time \\
\hline
upper&1 & \SI{3.52}{\meter\per\second} & \SI{126}{\meter} & \SI{35.8}{\second} \\
&2, 3 & \SI{3.01}{\meter\per\second} & \SI{138}{\meter} & \SI{45.8}{\second} \\
&4$\ldots$9 & \SI{3.04}{\meter\per\second} & \SI{189}{\meter} & \SI{62.2}{\second} \\ \hline
lower & 10 & \SI{3.80}{\meter\per\second} & \SI{220}{\meter} & \SI{57.9}{\second} \\
&11& \SI{3.09}{\meter\per\second} & \SI{134}{\meter} & \SI{43.4}{\second} \\
&12, 13, 14& \SI{3.10}{\meter\per\second} & \SI{139}{\meter} & \SI{44.8}{\second} \\ \hline
total &entire trail& \SI{3.26}{\meter\per\second} & \SI{946}{\meter} &  \\
 &upper part& \SI{3.15}{\meter\per\second} & & \\ 
 &lower part& \SI{3.37}{\meter\per\second} & &  \\ \hline
\end{tabular}
\label{tab:allmend_real}
\end{table}
As demonstrated in multimedia extensions~\ref{mov:method} and~\ref{mov:allmend_racing}, the entire trail was annotated as traversable, allowing local path selection to emerge naturally from the policy rather than being strictly dictated, a skill which we refer to as ``local navigation''. For instance, the robot chose to drive over the seesaw rather than taking an available bypass. This behavior likely stems from the training distribution, where simulated slopes typically span the full width of the track, biasing the agent to traverse rather than circumvent such features. Conversely, the robot tended to drive alongside skinnies instead of balancing on top of them. This aligns with our observation made in simulation, where the policy only reacts to skinnies upon front-wheel impact -- a limitation caused by the sparsity of the height scan, which renders narrow geometric features largely unobservable prior to contact. At tabletop drops, we imposed no target descent points; instead, the robot dynamically committed to a descent trajectory based on its approach angle. Finally, the policy was not forced onto the wall-ride, as the adjacent gravel path was also segmented as traversable. Because the policy learned to corner on both flat ground and banked surfaces during training, it fluidly transitioned between these two modes during deployment, dictated solely by its entry velocity at the obstacle.

\subsection{Comparison with a Hierarchical Baseline}\label{ssec:comparison_hierarchical}
To compare against a hierarchical alternative, we retrained the velocity-tracking policy of \citet{bokser2026umv} with our height-scan observation, keeping the remaining training setup, including the terrain generator, identical to \citet{bokser2026umv}. Commanded linear velocities ranged from \SIrange{0}{3}{\meter\per\second} and angular velocities up to \SI{1.5}{\radian\per\second} in magnitude. We trained the policy for 16{,}000 learning iterations, and refer to it as ``baseline''. For completeness, we also include the original blind locomotion policy of \citet{bokser2026umv}, which we call ``baseline$^{\dagger}$ ''. Each locomotion policy was paired with one of two trail-following controllers: one that steers toward the centroid of the trail-labeled scan points, and a pure-pursuit controller with a look-ahead distance of \SI{2}{\meter}. In both cases, the trail follower controller implemented a fixed heading velocity of \SI{1.5}{\meter\per\second} or \SI{3}{\meter\per\second}.

All policies, including ours, were evaluated on 850 terrain patches from the \ac{TL}, comprising 10 instances of each terrain type across five difficulty levels. We removed all trees to avoid \ac{OOD} observations for the perceptive baselines, which, if anything, favors the baseline. For each difficulty level, we simulated 4{,}000 robots with all domain randomizations disabled and recorded the survival rate, the time to reach the final platform, the time spent off the trail, and the average speed. The episode length was \SI{30}{\second} for the \SI{1.5}{\meter\per\second} target and \SI{16}{\second} for the \SI{3}{\meter\per\second} target. The results are summarized in Tab.~\ref{tab:policy_evaluation2}.

The centroid-based controller performed slightly better than pure pursuit, and the blind locomotion policy outperformed its perceptive counterpart. We attribute the latter to the difference in training terrains: the perceptive baseline receives height scans on the \ac{TL} terrains that lie outside its training distribution, whereas the blind policy does not depend on them. This underlines the importance of matching the training terrain to the deployment environment, which the \ac{TL} provides.

The decline in success rate at higher target velocities suggests that a competitive command interface would also need to modulate the linear velocity based on the local terrain. We therefore trained a separate navigation policy on top of the blind baseline, using the same trail terrains as for our generic racing policy. The navigation policy observes the same signals as the generic racing policy and outputs twist commands at \SI{50}{\hertz}, the update rate of the locomotion policy, with actions clipped at \SI{1.5}{\radian\per\second} and \SI{3}{\meter\per\second}. We sample heading velocity targets for the downstream locomotion policy uniformly and keep them constant within a window of \SIrange{2}{7}{\second}. The navigation policy is encouraged to output the same values as the heading target, but can deviate from it to maximize survival rate.
The locomotion policy remains fixed during training and is treated as part of the environment, i.e., no gradients are propagated through it. As reward, we use the exploration reward (Eq.~\ref{eq:r1}) with $v_\text{min} = \SI{1}{\meter\per\second}$, the survival reward (Eq.~\ref{eq:r2}), and a reward for twist tracking of the same shape as the track-stand reward (Eq.~\ref{eq:track_stand}) with a weight of $2$. Episodes terminate when the locomotion policy triggers a termination \citep{bokser2026umv} or when the agent reaches the final platform; as we are only interested in traversal, there is no standing phase.

With a learned navigation policy, the hierarchy became the most competitive baseline (Table~\ref{tab:policy_evaluation2}). The best such setup, baseline-3$^{\dagger\top}$, survived in \SI{85}{\percent} to \SI{46}{\percent} of the runs at a realized average velocity of \SI{2.2}{\meter\per\second}, whereas our generic policy survived in \SI{99}{\percent} to \SI{91}{\percent} of the runs at \SI{3.3}{\meter\per\second}. Despite driving about \SI{50}{\percent} faster, our policy thus failed roughly an order of magnitude less often (6 to 15 times, depending on the difficulty level), with the gap widening on the harder terrain levels. The baseline also spent substantially more time off the trail. Unlike the heuristic controllers, the learned navigation policy maintained its survival rate at the higher speed limit, presumably because it learned to modulate its speed.


\begin{table*}[htbp]
\centering
\caption{\textbf{Evaluation of the generic racing policy and comparison with hierarchical baselines.} The ``ours'' policy is the one deployed on the last Allmend deployment day; its heading base velocity, averaged over all terrain levels, is \SI{3.3}{\meter\per\second}. The baseline locomotion policies are conditioned on a twist command, generated either by a controller that steers toward the centroid of the observed trail (no symbol), by a pure-pursuit controller ($^*$), or by a learned navigation policy ($^\top$), with a heading target velocity of \SI{1.5}{\meter\per\second} (baseline-1.5) or \SI{3}{\meter\per\second} (baseline-3). The dagger ($^\dagger$) denotes the blind locomotion policy.}
\label{tab:policy_evaluation2}
\begin{tabular}{@{}lccccc@{}}
\toprule
& Level & Survival Rate & Time to Goal (s) & Time Not on Trail (s) & Avg Speed (m/s) \\ \midrule
ours & 0 & \textbf{0.99} & \textbf{8.82} & \textbf{0.013} & \textbf{3.29} \\
 & 1 & \textbf{0.98} & \textbf{8.88} & \textbf{0.015} & \textbf{3.30} \\
 & 2 & \textbf{0.98} & \textbf{9.00} & \textbf{0.011} & \textbf{3.30} \\
 & 3 & \textbf{0.95} & \textbf{9.11} & \textbf{0.013} & \textbf{3.33} \\
 & 4 & \textbf{0.91} & \textbf{9.61} & \textbf{0.038} & \textbf{3.30} \\ \midrule[\heavyrulewidth]
baseline-1.5 & 0 & 0.85 & 19.7 & 0.767 & 1.45 \\
& 1 & 0.74 & 19.9 & 1.380 & 1.46 \\
& 2 & 0.60 & 20.3 & 1.860 & 1.45 \\
& 3 & 0.41 & 20.3 & 2.780 & 1.48 \\
& 4 & 0.22 & 22.1 & 6.230 & 1.41 \\ \midrule
baseline-1.5$^*$ & 0 & 0.86 & 19.8 & 0.820 & 1.46 \\
 & 1 & 0.72 & 20.1 & 1.47 & 1.45 \\
 & 2 & 0.57 & 20.4 & 2.19 & 1.44 \\
 & 3 & 0.37 & 20.7 & 3.58 & 1.45 \\
& 4 & 0.20 & 22.6 & 7.43 & 1.38 \\ \midrule
baseline-1.5$^\dagger$ & 0 & 0.86 & 19.4 & 0.270 & 1.46 \\
 & 1 & 0.78 & 19.6 & 0.389 & 1.49 \\
 & 2 & 0.72 & 19.9 & 0.390 & 1.45 \\
 & 3 & 0.60 & 20.4 & 0.795 & 1.46 \\
 & 4 & 0.44 & 21.2 & 1.420 & 1.44 \\ \midrule
baseline-1.5$^{\dagger\top}$ & 0 & 0.83 & 22.3 & 0.162 & 1.28 \\
 & 1 & 0.78 & 22.4 & 0.334 & 1.29 \\
 & 2 & 0.71 & 22.2 & 0.341 & 1.32 \\
 & 3 & 0.64 & 22.3 & 0.350 & 1.34 \\
 & 4 & 0.45 & 22.6 & 0.620 & 1.38 \\ \midrule[\heavyrulewidth]
baseline-3 & 0 & 0.83 & 9.8 & 0.881 & 2.95 \\
 & 1 & 0.62 & 10.1 & 1.160 & 2.90 \\
 & 2 & 0.42 & 10.2 & 1.720 & 2.88 \\
 & 3 & 0.26 & 10.4 & 2.320 & 2.88 \\
 & 4 & 0.12 & 10.7 & 3.830 & 2.91 \\ \midrule
baseline-3$^*$ & 0 & 0.74 & 9.90 & 0.962 & 2.94 \\
 & 1 & 0.53 & 10.2 & 1.66 & 2.85 \\
 & 2 & 0.33 & 10.2 & 2.02 & 2.88 \\
 & 3 & 0.20 & 10.5 & 2.56 & 2.86 \\
 & 4 & 0.09 & 11.1 & 4.94 & 2.80 \\  \midrule
baseline-3$^\dagger$& 0 & 0.85 & 9.77 & 0.162 & 2.96 \\
 & 1 & 0.67 & 9.93 & 0.378 & 2.94 \\
 & 2 & 0.49 & 10.1 & 0.793 & 2.90 \\
 & 3 & 0.33 & 10.3 & 1.540 & 2.91 \\
 & 4 & 0.19 & 10.5 & 2.130 & 2.96 \\ \midrule
baseline-3$^{\dagger\top}$ & 0 & 0.85 & 13.1 & 0.133 & 2.18 \\
 & 1 & 0.79 & 12.3 & 0.210 & 2.35 \\
 & 2 & 0.72 & 12.9 & 0.212 & 2.28 \\
 & 3 & 0.58 & 14.2 & 0.255 & 2.12 \\
 & 4 & 0.46 & 14.7 & 0.488 & 2.11 \\ \bottomrule
\end{tabular}
\end{table*}

\subsection{Failure Case Analysis for Trail Deployment} \label{ssec:fail}
\begin{figure*}[h!]
\centering
\includegraphics[width=\textwidth]{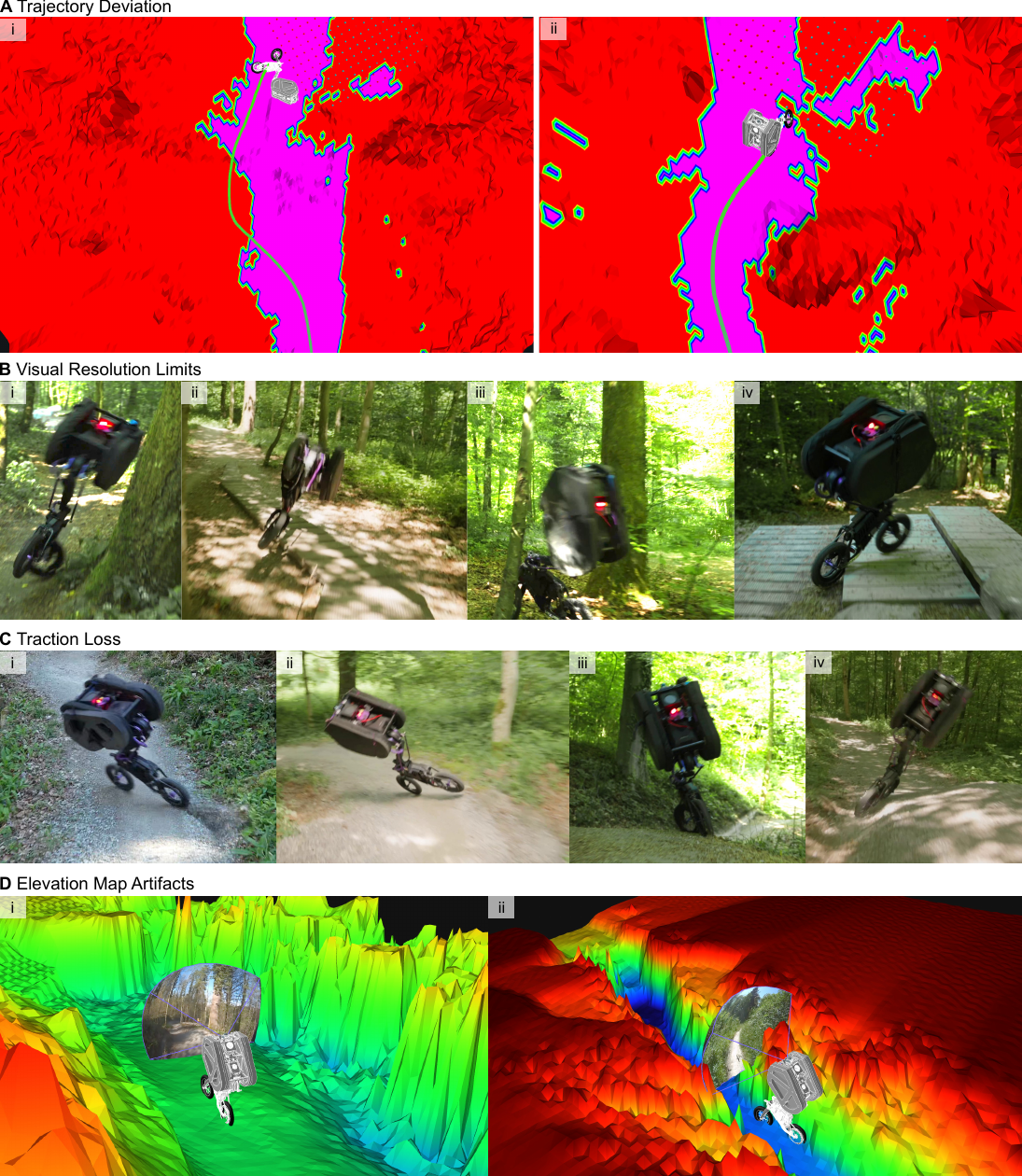}
\caption{\textbf{Primary failure modes during real-world deployment.} Each snapshot captures the state immediately preceding a critical loss of balance. \textbf{(A)} The robot exited the trail and failed to route back into the safe zone. The green path visualizes the base position. (i) The robot deviates from track boundaries and aggressively steers inwards, leading to slip. (ii) A wrongly perceived trail causes the robot to drive off the track. \textbf{(B)} The robot collided with an obstacle whose height or width falls below the spatial resolution of the elevation map. \textbf{(C)} The rear wheel entered a slip phase. \textbf{(D)} Trees are registered as untraversable walls in the elevation map (i), while tall grass narrows the traversable path (ii). Insets display the corresponding onboard camera perspective.}
\label{fig:failure}
\end{figure*}
We conducted 79 autonomous runs on four deployment days between early April and late May. We exclude the second day (April 24), on which testing was stopped after 10 runs, from the quantitative analysis; its failures are summarized in Appendix section~\ref{supp:deployment_days}. The following analysis covers the remaining 67 runs, excluding the two runs that ended in a track-stand. We then categorized the primary failure modes, which are exemplified in Fig.~\ref{fig:failure}.

The most frequent cause of critical balance loss (23/67) occurred when the robot deviated from the physical track (\ref{fig:failure}A). This trajectory deviation was generally caused either by the control policy failing to safely reject external disturbances (\ref{fig:failure}Ai), or by the semantic perception pipeline generating permissive traversability boundaries (\ref{fig:failure}Aii). In some scenarios (20/67), the robot suffered unrecoverable wheel slip on dry gravel (\ref{fig:failure}C), which happened during tight cornering (\ref{fig:failure}Ci, ii) or off-camber descents (\ref{fig:failure}Ciii, iv). Despite dedicated efforts in parameterizing ground friction and choosing conservative friction intervals (Appendix section~\ref{supp:dr}), the problem of wheel slip could not be completely resolved. Another prevalent failure mode (14/67) involved front-wheel or upper body collisions with narrow obstacles (\ref{fig:failure}B), such as roots (\ref{fig:failure}Bi), skinnies (\ref{fig:failure}Bii, iv), or thin tree trunks (\ref{fig:failure}Biii). These obstacles are too small to appear in the height scan and are mostly invisible to the policy.

Finally, perception artifacts occasionally (10/67) prevented the elevation mapping pipeline from reconstructing the geometry of the upcoming path (Fig.~\ref{fig:failure}D). We identified two primary mechanisms driving these mapping failures, both rooted in the 2.5D elevation map recording the maximum height within each grid cell. First, trees and dense woody vegetation are registered as tall, solid geometries, constructing untraversable walls across the path (Fig.~\ref{fig:failure}Di). Second, tall grass yields out-of-distribution height maps: although compliant and traversable, it is captured at its full height, which can narrow or even block the perceived corridor (Fig.~\ref{fig:failure}Dii). Both effects are aggravated at high speed, when the policy pitches the torso backward for dynamic stability and angles the onboard range sensor upward, reducing the density of ground returns, forcing the elevation map to extrapolate from sparse data. 

\section{Discussion} \label{sec:discussion}

\subsection{Performance Gain Through Specialization} \label{ssub:overfitting}
Our results are in line with recent legged parkour frameworks, which indicate that specializing control policies to known geometric topologies significantly enhances peak performance~\citep{rudin2022advanced}. When training the specialized racing policy, the average simulated speed reached \SI{5.57}{\meter\per\second} on the Allmend trail (subsection~\ref{ssec:simulation}). Comparing this to the generic real-world deployment on Allmend (\SI{3.26}{\meter\per\second}) (Tab.~\ref{tab:allmend_real}), the performance gain of geometric specialization corresponds to a roughly 1.7-fold increase in average velocity (simulated vs. real-world).

While the generic racing policy collides with the slalom poles, the specialized policy is capable of completing the same trail with a \SI{89}{\percent} (for $v_\text{min}=10$) to \SI{100}{\percent} (for $v_\text{min}=2$) success rate (subsection~\ref{ssec:simulation}). The specialized policies seem to remember the critical turns, likely based on surrounding landmarks. While this presents a potential solution to the problem of ``blind spots'', it also introduces a new weak link for deployment, as those geometric or semantic landmarks are not clearly defined nor guaranteed to persist over time.

\subsection{Sim-To-Real Gap} \label{ssub:sim2real_gap}
In simulation, the specialized racing policy with dedicated tuning reached an average speed of \SI{9.8}{\meter\per\second} on the upper Allmend section (subsection~\ref{ssec:simulation}), whereas real-world deployment velocities ranged from \SIrange{3.0}{3.8}{\meter\per\second} (subsection~\ref{ssec:trail_racing}). This discrepancy does not measure the sim-to-real gap, as the two results stem from different policies: the specialized policy was trained on a single \ac{DT} with the minimum exploration velocity raised to \SI{10}{\meter\per\second}, whereas the generic racing policy deployed in the real world was trained across \SI{46}{\kilo\meter} of randomized synthetic trails. The real-world speeds reflect a deliberate operating point rather than a hardware or simulation limit: we tuned the \ac{MDP} to trade off peak speed for the robustness required to survive unstructured terrain, sensor noise, and perception errors.

We therefore quantify the sim-to-real gap only where the same policy runs on the same track in simulation and on hardware. For control alone, the \ac{MoCap} parkour policy succeeded in \SI{99}{\percent} of simulated runs and in 8 out of 10 real runs (subsection~\ref{ssec:parkour}). For the full perception-and-control loop, the generic racing policy completed the indoor circuit in \SI{6.1}{\second} per lap in simulation and \SI{6.85}{\second} on hardware, corresponding to an \SI{11}{\percent} lower average speed (subsection~\ref{ssec:trail_indoor}). For the outdoor trail, no like-for-like comparison exists (subsection~\ref{ssec:trail_indoor}).

Deploying the specialized racing policy to the physical trail is outside of the scope of this work, as it would introduce several compounding challenges, revealing an unsolved sim-to-real gap at higher velocities. First, previous work identified the UMV's physical speed limit on flat indoor ground as \SI{8}{\meter\per\second}~\citep{bokser2026umv}, establishing a hardware ceiling below the simulated peak velocity. Second, the height scan provides a relatively small lookahead window of roughly 3 meters, and the information within is relatively sparse. Third, the specialized policy is inherently overfitted to the geometry of a single \ac{DT} (subsection~\ref{ssub:overfitting}). The simulated mesh was generated from point clouds stripped of vegetation, and vertex simplification smoothed out fine-grained geometric disturbances like small rocks and roots. Because the real-world semantic elevation map contains these rich details, a specialized policy relying on memorized geometry would likely fail.

\subsection{Intersections and Forks} \label{ssec:intersection}
When encountering trail intersections or forks, \ac{SemMap} segments multiple valid paths simultaneously. Because the control policy was not exposed to branching trajectories during training, it lacks an implicit mechanism for global routing. Empirically, we observe that the policy resolves this ambiguity by defaulting to the path that requires the least steering effort (e.g., multimedia extension~\ref{mov:allmend_racing} at 1:17). While this emergent behavior allows the system to smoothly clear simple intersections at random, it creates a vulnerability when the perception pipeline generates spurious semantic artifacts, which can inadvertently divert the robot into untraversable terrain (subsection~\ref{ssec:fail}). 

\subsection{Comparison to Autonomous Hiking}
Our work intersects closely with the field of autonomous hiking (sections~\ref{sec:introduction} and \ref{sec:related_work}). While our perception capabilities can be evaluated against existing methods, a direct comparison of the control policies is difficult. Specifically, prior work~\citep{miki2023bridging, mattamala2025wvn, Reed2025, Giusti2016trail} prioritized traversal success over time-optimal performance. Consequently, to the best of our knowledge, these studies do not report the explicit velocity metrics required for a quantitative comparison.


\subsection{Limitations and Future Work}
Our extensive real-world deployments also expose clear limitations of the current system. We group these limitations into four core categories.
\begin{itemize}
\item Field deployment revealed perception vulnerabilities caused by tall grass and overhanging vegetation being erroneously fused into the elevation layer. These geometric artifacts resulted in paths that are artificially narrowed or even blocked entirely by the appearance of ``phantom'' vegetation walls. Two parallel directions could resolve this: Filter vegetation from the point cloud before fusing, or remove the explicit height representation altogether by learning locomotion from RGB streams.
\item Exposed roots, loose stones, thin tree trunks, and slippery gravel frequently caused unrecoverable force impacts or traction loss. Relevant visual details are captured in neither the sparse depth nor the semantic layer, but are rather considered as external disturbances to the controller. Substituting the sparse height scan with rich, high-dimensional visual latent features represents a vital path forward to enable human-competitive trail racing. The hardware could further benefit from a roll joint, potentially improving reactive behaviors of the upper body.
\item While cross-terrain generalization ensures broad policy robustness, it also restricts the maximum achievable racing velocities. We hypothesize that future strategies will bridge this gap by deploying specialized policies trained for the precise geometric nuances of isolated trail segments.
\item As our work focuses explicitly on ``local'' navigation, a foundational bottleneck of our proposed framework is its implicit reliance on a unique and continuous path. Elevating this architecture to true cross-country navigation capabilities requires integrating sparse directional guidance, potentially sourced from topological maps.
\end{itemize}

\subsection{Lessons Learned} \label{ssec:lessons}

Beyond the quantitative results, the development of the system yielded several lessons that we expect to transfer to other agile platforms and tasks.

\begin{itemize}
    \item \textbf{Shape behavior through the environment, not the reward.} With fixed reward weights, the terrain mesh and its semantic annotation became the main design tool: a beam placed in a gap, for instance, turned a rolling solution into an active jump (subsection~\ref{ssec:tean}). 
    \item \textbf{Generalization costs speed.} Policies trained on diverse procedural trails transferred robustly but raced well below the speeds of policies specialized to a single \ac{DT}, which in turn rely on landmarks that are not guaranteed to persist (subsection~\ref{ssub:overfitting}).
    \item \textbf{At speed, perception reliability matters as much as accuracy.} Without confidence-weighted fusion, low-confidence observations during fast turns progressively corrupted the map (Fig.~\ref{fig:indoor}D), and visual appearance shifted noticeably within a single season, so cheap in-field adaptation proved essential (subsection~\ref{ssec:semmap}).
    \item \textbf{Observability, not dynamics, limited the real-world performance.} Most failures stemmed from features the policy could not perceive, such as thin obstacles below the resolution of the height scan and vegetation registered as solid geometry (subsection~\ref{ssec:fail}).
    \item \textbf{Close one gap at a time.} Validating control, perception, and their integration separately, from \ac{MoCap} parkour to indoor racing, allowed us to attribute failures to a single component before deploying the full system outdoors.
\end{itemize}

\section{Conclusion}
In this work, we successfully tackled the challenge of autonomous trail racing based on the two core contributions, \ac{TENa} and \ac{SemMap}. 
Rather than serving as a domain-specific solution, \ac{TENa} is a unified \ac{MDP} formulation that scales across racing and parkour. Because the task objective is encoded in the terrain, the task reward and its weights are shared across all tasks, and behaviors can be shaped solely through terrain semantics and episode length. Furthermore, by collapsing the planning-and-tracking hierarchy into a single learning framework, \ac{TENa} reduces deployment complexity. 

Because the task objective is embedded in the terrain, navigation simplifies to following the perceived trail. Leveraging a frozen backbone and a lightweight dual-head decoder, \ac{SemMap} produces such a semantic 2.5D map at~\SI{30}{\hertz} (subsection~\ref{ssec:perception}). Because the map aggregates observations over time and a learned confidence signal down-weights unreliable predictions, transient perception errors are absorbed rather than propagated (section~\ref{supp:fusion}). As the decoder is lightweight and environment-specific, adaptation to a new deployment requires minutes of training on data that can be collected without a robot.

While our framework is platform-agnostic (Fig.~\ref{fig:parkour_generalization} and \ref{fig:teaser-g1}), we validated it on a statically unstable bicycle on a kilometer-scale mountain bike trail. The technical complexity of this endeavor was twofold. First, unlike stable quadrupeds or ground rovers, the \ac{UMV} must constantly balance. Second, the outdoor trail is subject to seasonal changes, leading to fluctuations in the visual appearance and terrain friction. To systematically mitigate these sim-to-real risks, we employed a progressive validation scheme:
\begin{enumerate}
\item Using two different robotic platforms, the \ac{UMV} and ANYmal, and 19 different tasks (17 parkour tracks, a~\SI{419}{\meter} long trail, and over~\SI{46}{\kilo\meter} of generic trails from the \ac{TL}), we demonstrated that agile skills emerge naturally as a function of the environment's geometry and semantics, without the need for task-specific reward tuning.
\item We isolated the control sim-to-real gap by deploying a locomotion policy in a \ac{MoCap}-instrumented parkour, achieving an~\SI{80}{\percent} success rate on a track comprising a slalom, a ramp jump, and a staircase landing. 
\item We evaluated \ac{SemMap} in isolation, demonstrating robust onboard reconstruction of terrain semantics, label-efficient adaptation competitive with fully fine-tuned baselines, and confidence-weighted fusion for reliable sim-to-real deployment. 
\item We closed the perception-and-control loop by deploying a generic racing policy to autonomously navigate a narrow, twisted racing track, where it completed 130 continuous laps at an average speed of~\SI{3.5}{\meter\per\second}. We further quantified the sim-to-real gap by comparing against simulated runs.
\item We demonstrated full onboard autonomy by traversing the complete kilometer-long Allmend trail within six runs, reaching average segment velocities of up to~\SI{3.8}{\meter\per\second} and peak velocities larger than~\SI{5}{\meter\per\second}. We compared the results to a teleoperated locomotion policy and showed an increase in average velocity of a factor of 2.3 on the upper Allmend trail. In simulation, our generic policy failed up to an order of magnitude less often than hierarchical baselines built on the policy of \cite{bokser2026umv}, including one with a navigation policy learned on the same trail terrains, while driving faster.
\end{enumerate}

\theendnotes

\begin{acks}
The authors would like to thank Michael Riner and Per Frivik for their assistance during field deployment; Marco Zenti for hardware support; Diego Ferigo for his contributions to the deployment code; Petr Listov and Adrian Brandemuehl for their help with the \ac{MoCap} system; Andrea Sgobbi for assistance with sensor integration; James Tigue for assistance with the simulation and training code; and Tristan Laidlow for feedback on perception. Lastly, the authors extend their gratitude to the entire UMV and FIELD teams for continuous support across hardware and software.
\end{acks}

\begin{funding}
This research is supported by the Robotics and AI (RAI) Institute, which received funding from the Hyundai Motor Group. The opinions expressed herein are those of the authors and do not necessarily reflect the views of the funding organizations.
\end{funding}

\begin{dci}
The authors declare that they have no competing financial or personal interests that could influence the work reported in this paper.
\end{dci}

\begin{sm}
The supplementary datasets, ROS bags from the Allmend trail deployment, and plotting scripts that support the findings of this study are openly available in Zenodo (https://doi.org/10.5281/zenodo.23063816). Specifically, the repository contains:
\begin{enumerate}[label=data S\arabic*:, ref=S\arabic*, align=left, leftmargin=*]
\item \label{supmat:dataS1} (dataS1.zip) Tabulated datasets used to generate Figures \ref{fig:label-efficiency}, \ref{fig:training-curve}, \ref{fig:ood-study}, and \ref{fig:inference-study}.
\item \label{supmat:rosbags} (rosbag\_trail1.zip,\ldots. rosbag\_trail6.zip) ROS bags recorded during the Allmend trail deployment, containing the raw telemetry and perception data used to create the values in Tab.~\ref{tab:allmend_real}.
\item \label{supmat:foxglove} (foxglove\_layout.json) Foxglove configuration layout used to visualize the provided ROS bags.
\item \label{supmat:indoor} (Figure\_7C.zip) Raw data and the corresponding Python script required to reproduce the plots in Fig.~\ref{fig:indoor}C.
\item \label{supmat: friction} (plot\_friction\_Fig\_S1.py) Python script used to generate the friction plot shown in Fig.~\ref{fig:friction}.
\end{enumerate}
\end{sm}

\bibliographystyle{SageH}
\bibliography{Sage.bib}

\begin{appendix}
\setcounter{table}{0}
\setcounter{figure}{0}
\renewcommand{\thetable}{S\arabic{table}}
\renewcommand{\thefigure}{S\arabic{figure}}

\section{Index to Multimedia Extensions}
The following videos can be found alongside the publication
\begin{enumerate}[label=Multimedia Ext. E\arabic*:, ref=E\arabic*, align=left, leftmargin=*]
\item \label{mov:method} narrated overview of the method and results
\item \label{mov:allmend_racing} autonomous racing on Allmend trail
\item \label{mov:sim_parkour} simulated parkour with \ac{UMV} 
\item \label{mov:sim_allmend} simulated Allmend trail with \ac{UMV}
\item \label{mov:mocap_parkour} \ac{MoCap} assisted indoor parkour
\item \label{mov:perceptive_indoor_racing} perceptive indoor trail racing
\item \label{mov:sim_parkour_anymal} simulated parkour with ANYmal
\item \label{mov:sim_allmend_anymal} simulated Allmend trail with ANYmal
\item \label{mov:semmap} \ac{SemMap} perception system overview
\end{enumerate}

\section{Nomenclature}
\acrodefplural{DOF}[DOFs]{Degrees of Freedom}
\begin{acronym}[SemMap]
    \acro{RL}{reinforcement learning}
    \acro{PPO}{Proximal Policy Optimization}
    \acro{UMV}{Ultra Mobility Vehicle}
    \acro{MPCC}{Model Predictive Contouring Control}
    \acro{MPC}{Model Predictive Control}
    \acro{MLP}{Multi-Layer Perceptron}
    \acro{AMP}{Adversarial Motion Prior}
    \acro{MDP}{Markov decision process}
    \acro{TL}{Trail-Library}
    \acro{SemMap}{Semantic Maps}
    \acro{MoCap}{Motion Capture System}
    \acro{VFM}{Vision Foundation Model}
    \acro{OOD}{out-of-distribution}
    \acro{mIoU}{mean Intersection over Union}
    \acro{ViT}{Vision Transformer}
    \acro{PCA}{Principal Component Analysis}
    \acro{MPS}{Multi-Process Service}
    \acro{GIL}{Global Interpreter Lock}
    \acro{FPS}{frames per second}
    \acro{VLM}{vision-language model}
    \acro{BEV}{bird's-eye view}
    \acro{LIO}{LiDAR-inertial odometry}
    \acro{MEM}{multi-modal elevation mapping}
    \acro{TENa}{Terrain Embedded Navigation}
    \acro{DT}{Digital-Twin}
    \acro{DOF}{Degree of Freedom}
\end{acronym}

\section{Optimizer Settings} \label{supp:optimizer_settings}
\begin{table*}
\caption{\textbf{Optimizer settings}. \ac{PPO} hyperparameter configurations. Here, ``indoor'' denotes the short parkour circuits, whereas ``outdoor'' refers to outdoor trails spanning several hundred meters in length. Numbers provided in curly brackets indicate task dependency.}
\centering
\begin{tabular}{lccc}
\hline
Setting &  \ac{TL} & DT indoor & DT outdoor \\
\hline
Number of environments & $4096$ & $4096$ & $1024$ \\
Number of steps & $24$ & $24$ & $192$ \\
Episode length in seconds & $16$ & $\{8, 10, 12, 14\}$ & $120$ \\
Number of epochs & $16000$  & $16000$ & $16000$ \\
Entropy coefficient & $0.005$  & $0.008$ & $0.008$ \\ \hline
\end{tabular}
\label{tab:ppo}
\end{table*}
We optimize the unified locomotion and navigation policies using \ac{PPO}~\citep{schulman2017ppo} and the hyperparameters specified in Table~\ref{tab:ppo}. For the extended outdoor trails, we increase the rollout duration to properly capture the longer task horizons. All remaining, unspecified hyperparameters are kept identical to~\cite{rudin2021learning}.

\section{Simulator Settings} \label{supp:sim}
We employ the compliant contact model provided by Isaac Sim, utilizing the specific rigid-body and contact parameters previously identified in~\cite{bokser2026umv}. Furthermore, we approximate the tire geometry using a mathematical torus primitive rather than a standard meshed cylinder. Empirically, this representation yielded more stable behaviors during tight turning scenarios. We note, however, that computing collisions between torus primitives and dense terrain meshes decreases simulation rollout speeds by approximately \SI{20}{\percent} compared to a simplified cylinder mesh. Similar to \cite{bokser2026umv}, the control policy is queried at~\SI{50}{\hertz}, but we update the physics simulation at a frequency of~\SI{500}{\hertz} as opposed to \SI{200}{\hertz}.

We observed that enforcing explicit linear or angular velocity limits on articulations within Isaac Sim causes the physics engine to inject artificial external forces to satisfy those constraints. To prevent the policy from exploiting these artificial forces, we explicitly disable these constraints.

\section{Network Architecture and Observations} \label{supp:observations}
We employ an asymmetric actor-critic framework in which the policy (actor) and the value function (critic) are parameterized by separate \acp{MLP}, similar to \cite{bokser2026umv} but using larger network sizes. The actor network models a stochastic policy $\pi(\vec{a} \mid \vec{o}) \sim \mathcal{N}(\vec{\mu}_\theta(\vec{o}), \vec{\sigma}_\theta)$, where the action mean $\vec{\mu}_\theta(\vec{o})$ is mapped by an \ac{MLP} parameterized by $\theta$, and the log-standard deviation $\vec{\sigma}_\theta$, with the latter being a learned state-independent layer. Conversely, the critic network is represented by a deterministic value function $V_\phi(\vec{o}, \vec{\tilde{o}})$ parameterized by $\phi$. To accelerate policy convergence during simulation, the critic is granted access to both the standard observation vector $\vec{o}$ and privileged state vector $\vec{\tilde{o}}$. Both the actor and critic networks feature three hidden layers comprising 512 neurons each. To stabilize the learning process, all incoming observation streams are dynamically normalized online utilizing running means and standard deviations. A comprehensive breakdown of all observation terms, alongside the specific noise levels applied to them during training, is detailed in Table~\ref{tab:observations}. The noise distributions are mostly similar to \cite{bokser2026umv}, however, we have slightly tuned the specific shape and parameters to balance our reward design. 
\begin{table*}[tbp]
\caption{\textbf{Observation space}. The critic network extends the actor's inputs with privileged, noise-free observations. The scan pattern consists of a $16 \times 16$ grid with a spatial resolution of \SI{0.2}{\meter}, centered \SI{2}{\meter} ahead of the robot's base frame. The notation $d(0.05)$ refers to positional drift, which is applied independently across the $x$, $y$, and $z$ axes. The semantic observation layer incorporates a custom label-corruption scheme. Finally, the IMU observation vector comprises six dimensions, sourced from two distinct IMU sensors: one mounted on the bike's base and the other on the upper body.
}
\centering
\setlength{\tabcolsep}{5pt}
\begin{tabular}{llcc}
 \hline
Type & Observations & Dim & Noise \\ \hline
Actor & Fork angle & 1 & $\mathcal{N}(\mu=0, \sigma=0.001)$ \\
& Joint body angles & 2 & $\mathcal{U}(\pm 0.01)$ \\
& Fork velocity & 1 & $\mathcal{N}(\mu=0, \sigma=0.1)$ \\
& Rear-wheel velocity & 1 & $\mathcal{N}(\mu=0, \sigma=0.05)$ \\
& Joint body velocities & 2 & $\mathcal{U}(\pm 0.15)$ \\
& IMU angular velocity & 6 & $\mathcal{U}(\pm 0.1)$ \\
& IMU projected gravity & 6 & $\mathcal{U}(\pm 0.015)$ \\
& Standing command & 1 & $0$ \\
& Scan, height layer & 256 & $\mathcal{U}(\pm 0.08)$ and $d(0.05)$ \\
& Scan, semantic layer & 256 & custom \\
& Last action & 4 & $0$ \\ \hline
Critic & base linear velocity & 3 &  \\
& Base pose & 6 &  \\
& Time left in the episode & 1 &  \\  \hline
\end{tabular}
\label{tab:observations}
\end{table*}

During real-world deployment, \ac{SemMap} may produce incorrect semantic annotations. To expose the policy to these artifacts during training, we introduce three distinct noise models. First, the label of any individual scan point is randomly flipped with a probability of~\SI{4}{\percent}. Second, there is a~\SI{20}{\percent} probability that a random block of consecutive rows in the scan pattern is entirely misclassified as non-trail. Third, we flip the labels of valid trail points with a probability that increases linearly with the scan row index. Specifically, the corruption probability is set to~\SI{0}{\percent} for the scan row closest to the robot and scales up to~\SI{20}{\percent} for the furthest row, with the probabilities for intermediate rows interpolated linearly.

The three noise models were designed to mimic error patterns we observed in \ac{SemMap} outputs during development: isolated label flips at class boundaries, contiguous unsegmented regions caused by occlusions, and decreasing segmentation reliability with distance. Their parameters were set heuristically and calibrated during the indoor racing experiments (Subsection~\ref{ssec:trail_indoor}). We did not perform a systematic ablation of the individual noise terms; policies trained with the full model transferred to both the indoor and outdoor deployments without further modification.

\section{Action Space and Low-Level Control} \label{supp:actions}
The action space is identical to \cite{bokser2026umv} and is repeated here for completeness.
The neural network predicts positional targets for the steering fork, lower-body joints, and upper-body joints, alongside a target angular velocity for the rear wheel. The front wheel remains unactuated. The high-level policy actions are updated at \SI{50}{\hertz}, while a low-level PD controller converts these targets into joint torques at \SI{8}{\kilo\hertz}. For the position-targeted joints, the applied torque $\tau_i$ is computed as $\tau_i = k_{p,i}(q_{i,\text{des}} - q_i) - k_{d,i}\dot{q}_i$, whereas for the velocity-targeted joint, the torque is computed using a proportional velocity error $\tau_i = k_{p,i}(\dot{q}_{i,\text{des}} - \dot{q}_i)$. A comprehensive summary of all action scales and joint-specific PD control gains is provided in Table~\ref{tab:actions}.
\begin{table}[tbp]
\caption{\textbf{Action space and low-level control parameters.} The raw, normalized outputs of the actor network are multiplied by their respective action scales to generate the desired position ($q_{i,\text{des}}$) or velocity ($\dot{q}_{i,\text{des}}$) targets. The low-level PD controller tracks these targets with gains $k_p$ and $k_d$. The front wheel is unactuated and operates passively.}
\centering
\begin{tabular}{l c c c c c}
\hline
Joint & Dim & Type & Action Scale & $k_p$ & $k_d$ \\ \hline
Fork & 1 & pos & $1.0$ & $1.4$ & $0.6$ \\
Body & 2 & pos & $1.5$ & $140$ & $10$ \\
Rear wheel & 1 & vel & $15.0$ & $0.6$ & $0$ \\
Front wheel & 1 & -- & -- & -- & -- \\ \hline
\end{tabular}
\label{tab:actions}
\end{table}

\section{Style Rewards} \label{supp:style}
We compute each style reward as $r_i = w_i \cdot f_i(e_i)$, where $w_i$ denotes the respective scalar weight, and $f_i$ represents the mathematical shaping function that acts upon an internal state or control error $e_i$. These style rewards do not define the task itself, but are necessary for regularizing the policy's outputs to ensure safe and smooth sim-to-real transfer. The complete formulation of all style rewards is detailed in Table~\ref{tab:style}. We note that we use substantially fewer style rewards compared to \cite{bokser2026umv} to not counteract the emergence of high-speed and agile behaviors. In particular, we do not constrain the upper body joints to track a default joint position, but let the body rather discover optimal configurations on its own. It needs to be said, however, that the choice of default joint positions nevertheless influences the overall behavior, even if they are not explicitly tracked.
\begin{table}[tbp]
\caption{\textbf{Style rewards}. Summary of the error signals $e_i$, shaping functions $f(e_i)$, and respective weights $w_i$. Here, $s_i$ represents the action scale of the corresponding joint; $a_k$ and $a_{k-1}$ denote the raw network actions at the current and previous time steps; $\dot{q}_a$ are the measured angular velocities; $\tau_{\text{des}}$ represents the desired joint torques after clipping.}
\centering
\begin{tabular}{lccc}
 \hline
reward & $e$ & $f(e)$ & $w_i$ \\ \hline
Fork action & $s_f \cdot (a_k - a_{k-1})$ & $\sum |e| + e^2 $ & $-1e^{-4}$ \\
Wheel action & $s_w \cdot (a_k - a_{k-1})$ & $\sum |e| + e^2 $ & $-5e^{-6}$ \\
Body action & $s_b \cdot (a_k - a_{k-1})$ & $\sum |e| + e^2 $ & $-1e^{-3}$ \\
Joint vel & $\dot q_a$ & $\sum |e| + e^2 $ & $-1e^{-3}$ \\
Joint power & $\dot q_a \cdot \tau_\text{des}$ & $\sum |e| + e^2 $ & $-1e^{-7}$ \\ \hline
\end{tabular}
\label{tab:style}
\end{table}

\section{Domain Randomizations} \label{supp:dr}
\begin{table*}[tbp]
\caption{\textbf{Domain randomization parameters.} Summary of the randomization events, parameter bounds, and application modes utilized during training to facilitate sim-to-real transfer. For evaluation purposes, all events are disabled, and rigid body material friction is set to the average value of the boundaries.}
\centering
\begin{tabular}{l l c c}
\hline
Event & Parameter & Range & Mode \\ \hline
Randomize rigid body material & static friction & $[0.5, 1.2]$ & Startup \\
 & dynamic friction & $[0.5, 1.2]$ & Startup \\ \hline
Reset root state uniform & pose.z & $[0.0, 0.2]$ & Reset \\
(in base frame) & pose.roll & $[-0.04, 0.04]$ & Startup \\
 & twist.x & $[0.0, 2.0]$ & Startup \\ \hline
Randomize rigid body inertia & relative mass & $[0.9, 1.1]$ & Startup \\ 
 & com.x, com.y, com.z & $[-0.03, 0.03]$ & Startup \\ \hline
Push by adding velocity & twist.x, twist.y & $[-0.6, 0.6]$ & Interval $[2, 12]$ \\ 
(in base frame) & & & \\ \hline
Apply external force/torque & absolute force & $[-2.0, 2.0]$ & Reset \\ \hline
Randomize joint parameters & absolute friction & $[0.0, 0.03]$ & Reset \\ \hline
\end{tabular}
\label{tab:dr}
\end{table*}
We employ standard domain randomization techniques provided by Isaac Lab to bridge the sim-to-real gap. The specific randomization events, their associated parameter ranges, and their application modes are summarized in Table~\ref{tab:dr}. We note that while the number of randomized parameters is substantially smaller than in~\cite{bokser2026umv}, we employ wider interval ranges for those selected parameters. We found that this approach yields a more tractable tuning problem, facilitating faster iteration and refinement. We also observed that the base twist randomization at startup is a particularly useful event for our task; it allows the agents to discover a stable forward motion almost instantaneously. Heuristically, we found that in our case randomizing actuator gains had no clear benefits and did not positively reflect in real-world success rates. We also found that the policies evaluated in simulation result in roughly similar performance metrics (slightly smaller average velocities without domain randomization), regardless of whether they were trained with or without domain randomizations (Appendix~\ref{supp:track_rand}).

\begin{figure}
\centering
\includegraphics{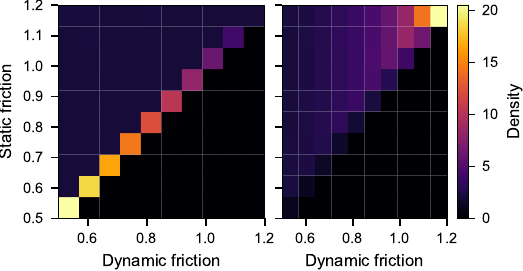}
\caption{\textbf{Comparison of friction coefficient sampling methods.} Both static and dynamic friction bounds are initialized to $\{0.5, 1.2\}$. We plot 1000 sampled friction pairs for two distinct strategies. Left: The default Isaac Lab approach, causing boundary accumulation. Right: Our adapted method, resulting in a more uniform distribution. The density values indicate the number of overlapping samples.}
\label{fig:friction}
\end{figure}

By default, the standard friction randomization event independently samples both static and dynamic friction coefficients from their specified ranges, and subsequently clips the dynamic friction to satisfy the physical constraint $\mu_{\text{static}} \ge \mu_{\text{dyn}}$. However, this clipping causes a disproportionate number of samples to accumulate directly on the boundary where static and dynamic friction equalize. To mitigate this boundary accumulation, we implement a modified sampling strategy: we first uniformly sample the dynamic friction coefficient, dynamically adjust the lower bound of the static friction range to match this sampled value, and finally sample the static friction from this constrained range. As illustrated in Fig.~\ref{fig:friction}, this approach yields a more uniform coverage of the feasible parameter space. While an alternative solution would be to sample uniformly across the entire valid triangular region, this would inadvertently skew the expected value of the dynamic friction toward the lower end of the spectrum.

\section{Cross embodied Generalization of \texorpdfstring{\ac{TENa}}{TENa} in Simulation} \label{supp:anymal}
\begin{figure*}[h!]
\begin{minipage}[t]{\dimexpr\textwidth-5.2in\relax}
\vspace{0pt}
\caption{\textbf{Generalization to quadrupedal parkour and trail running.} Snapshots of the ANYmal-D robot navigating \ac{DT} environments, executing jumping, climbing, and wall-run skills. All semantic meshes and the core \ac{MDP} components remained for all tasks strictly identical to those used for the wheeled robot. The episode length was the sole tuning parameter adjusted to account for the quadruped's different progression velocity.}
\label{fig:parkour_generalization}
\end{minipage}%
\hfill
%
\begin{minipage}[t]{5in}
\vspace{0pt} 
\includegraphics[width=5in]{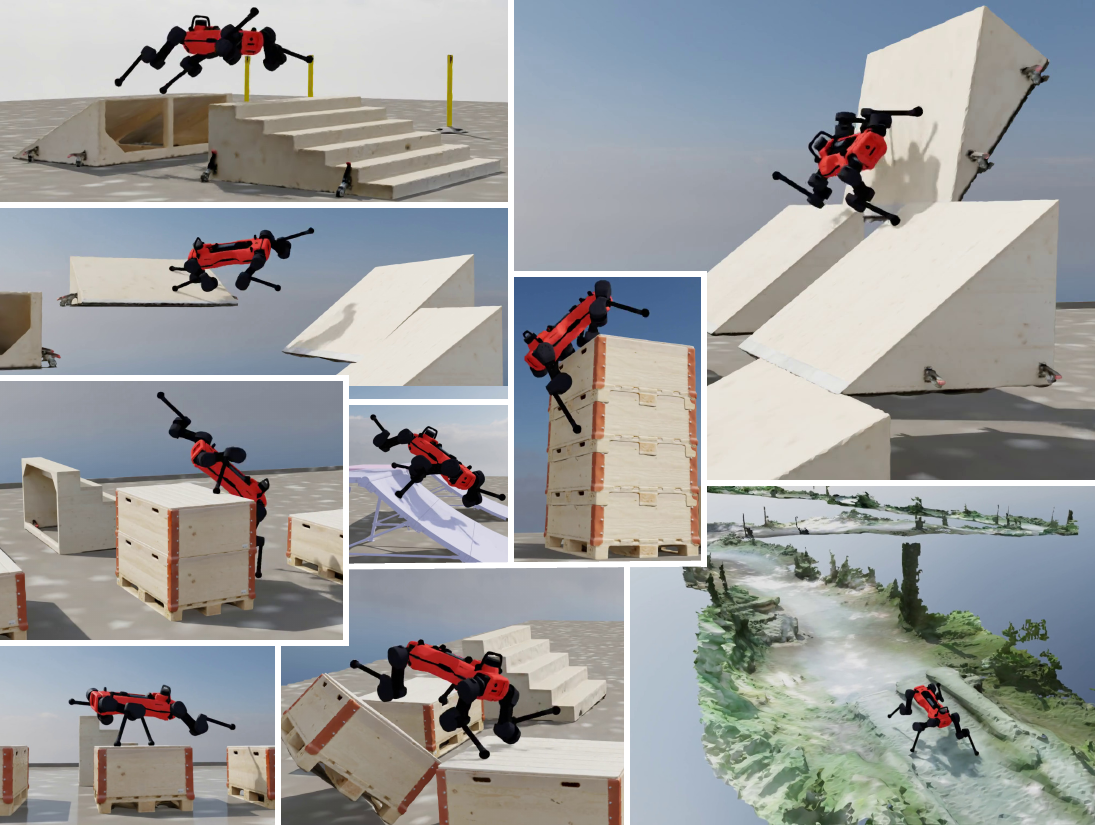}
\end{minipage}
\end{figure*}
While the \ac{UMV} serves as our primary evaluation platform, \ac{TENa} is platform-agnostic. To demonstrate robust learning generalization across different kinematic morphologies, we applied our \ac{MDP} formulation to an ANYmal-D quadruped robot. Out of the 12 parkour environments deployed in Fig.~\ref{fig:sim}A and Fig.~\ref{fig:indoor}A combined, ANYmal successfully navigated the maximum difficulty level of 8 terrains with a success rate exceeding~\SI{95}{\percent}, averaged over 1000 runs. For the remaining four terrains, the policy converged before mastering the final difficulty level, mostly due to foothold locations being too sparse. Furthermore, ANYmal traversed the upper part of the Allmend trail with a success rate of~\SI{81}{\percent} and within~\SI{89.5}{\second}, corresponding to an average speed of~\SI{4.68}{\meter\per\second}. Fig.~\ref{fig:parkour_generalization} highlights visual snapshots of all successful traversals, multimedia extension~\ref{mov:sim_parkour_anymal} shows the corresponding full parkour runs, and multimedia extension~\ref{mov:sim_allmend_anymal} demonstrates the ANYmal traversing the Allmend trail.

To interface the ANYmal-D quadruped within our simulation framework, we adapted the standard joint-level proprioceptive observations, action scales, default joint angles, and style rewards provided by the Isaac Lab quadruped environment. We slightly modified the observation space by replacing the velocity commands with our track-stand command and the height scan with the semantically annotated height scan. All core components of the \ac{MDP} formulation were transferred identically from the \ac{UMV} learning environment, with the episode length being the sole tuning parameter adjusted to account for the different morphological architecture. 

We note that we have not put any effort into style reward tuning and do not expect these policies to transfer zero-shot to the real hardware. Ill-tuned behaviors that likely will not transfer can be seen in both multimedia extensions. In~\ref{mov:sim_parkour_anymal} ANYmal uses its knees aggressively to climb over objects, and \ref{mov:sim_allmend_anymal} reveals an asymmetric running gait with the right-front leg being locked in.

\section{Perception Datasets} \label{supp:datasets}
Table~\ref{tab:datasets} lists the datasets used in the perception evaluation. The Grand Tour evaluation uses the left frames of the ZED2i stereo camera from the TRIM-1 sequence.

\begin{table*}[t]
\centering
\caption{\textbf{Datasets used in the perception evaluation.} The Lost and Found evaluation uses the 1{,}111 of 1{,}203 test images that contain obstacles.}
\label{tab:datasets}
\setlength{\tabcolsep}{4pt}
\small
\begin{tabular}{llcl}
\hline
Dataset & Classes & Train / Eval & Purpose \\
\hline
RELLIS-3D~\cite{jiang2021rellis3d} & 20$\rightarrow$3 & 3{,}302 / 983 & Label efficiency \\
Cityscapes~\cite{cordts2016cityscapes} & 19$\rightarrow$3 & 2{,}975 / -- & \acs{OOD} confidence \\
Lost and Found~\cite{pinggera2016lostandfound} & -- & -- / 1{,}203 & \acs{OOD} evaluation \\
ORFD~\cite{min2022orfd} & 3 & 8{,}392 / 2{,}193 & Inference quality \\
Indoor trail (ours) & 3 & 57 / 15 & Deployment (bicycle) \\
Outdoor trail (ours) & 5 & 28{,}763 / 3{,}196 & Deployment (bicycle) \\
Factory floor (ours) & 3 & 80 / 20 & Deployment (G1) \\
Grand Tour~\cite{frey2026grandtour} & -- & -- / eval & Generalization \\
\hline
\end{tabular}
\end{table*}

\section{Statistical Analysis of Perception} \label{supp:stats}
All perception comparisons follow a common protocol. The per-image studies (the ORFD upsampler study and the Lost-and-Found \ac{OOD} study) use paired Wilcoxon signed-rank tests on per-image metrics, with effect sizes given as the matched-pairs rank-biserial correlation $r$ and $P$-values Holm-corrected within each family of comparisons; because $n$ is large, we report effect sizes and magnitudes alongside $P$. Seed-based studies (label efficiency and training curves) report the mean $\pm$ standard deviation over $n=3$ independent training seeds that vary the random initialization and augmentation order while keeping the $k$-means label subsets fixed. The rapid-adaptation runs on the Grand Tour sites (Table~\ref{tab:grandtour-adaptation}) are single training runs per site and are reported as such, without variability estimates. Inference latencies report the mean $\pm$ standard deviation over $n=100$ timed iterations of a single pre-loaded image after 50 warm-up iterations on the AGX Orin, excluding transport overheads. Unless noted otherwise, the stated variability measure is the standard deviation. In all plots, error bars and shaded bands denote $\pm1$ SD; where they are not visible, the deviation is smaller than the marker or line width. Success rates of the real-world and simulated control experiments are reported as counts over the stated number of trials.

\section{Label Efficiency} \label{supp:label_efficiency}
We evaluate label efficiency on RELLIS-3D~\citep{jiang2021rellis3d}, an off-road dataset with 20 semantic classes widely used in traversability work. We compare \ac{SemMap} (frozen DINOv3-s, JAFAR upsampling, ${\sim}$100K trainable parameters) against DeepLabV3+~\citep{chen2018deeplabv3plus} with a COCO/VOC-pretrained ResNet-50~\citep{he2016resnet} (16.1M / 42.0M trainable parameters for decoder-only / end-to-end) and SegFormer-B0~\citep{xie2021segformer} with an ADE20K-pretrained MiT-B0 (400K / 3.7M), each with both a frozen backbone and full fine-tuning. Training images are selected with the FAISS-based pipeline of the main text, and we report the best validation \ac{mIoU} (983 images) with early stopping (patience 5). We additionally collapse the 20 classes into a 3-class traversability grouping (traversable, non-traversable, background).

\begin{figure}[tbp]
\centering
\includegraphics{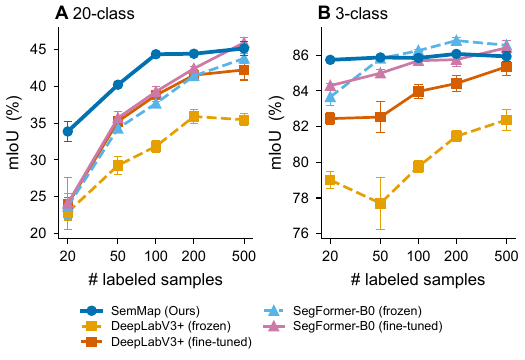}
\caption{\textbf{Label efficiency on RELLIS-3D} at 20 and 3 class granularities. Best validation \ac{mIoU} versus the number of labeled images (mean $\pm$ SD over $n=3$ seeds; bands smaller than the markers are not visible). \ac{SemMap} outperforms frozen and fine-tuned DeepLabV3+ and SegFormer-B0 at low label budgets, most clearly at 20 classes.}
\label{fig:label-efficiency}
\end{figure}

\begin{figure}[tbp]
\centering
\includegraphics{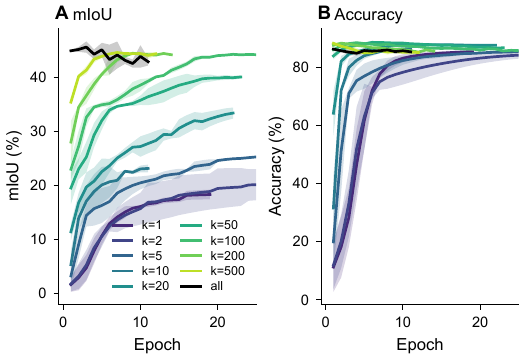}
\caption{\textbf{Validation \ac{mIoU} and accuracy vs.\ epoch} on RELLIS-3D (20 classes) at label budget $k$; bands show mean $\pm$ SD over $n=3$ seeds. With $k\geq100$ ($3\%$ of the training set), \ac{mIoU} matches the full dataset; accuracy converges within 3 epochs for $k\geq5$.}
\label{fig:training-curve}
\end{figure}

At 20 classes, \ac{SemMap} outperforms all baselines at low label budgets: $33.9 \pm 1.4\%$ \ac{mIoU} at 20 labeled images (mean $\pm$ SD over $n=3$ seeds) versus $23.9 \pm 0.9\%$ for fine-tuned DeepLabV3+ and $24.1 \pm 3.5\%$ for SegFormer-B0 (Fig.~\ref{fig:label-efficiency}). The low absolute \ac{mIoU} reflects severe class imbalance in RELLIS-3D, with pixel accuracy above $88\%$ even at the lowest budgets. Fully fine-tuned baselines need $2$--$5\times$ more labels to approach \ac{SemMap}, and the frozen ResNet-50 is weakest, confirming that ImageNet-supervised features are insufficient when the backbone cannot adapt. At 3 classes, the gap narrows and \ac{SemMap} plateaus from 20 samples, as every label covers all classes. Fig.~\ref{fig:training-curve} shows that with $k\geq5$ images, accuracy reaches ${\sim}80\%$ within 3 epochs, and that $k=100$ (3\% of the data) already matches the full training set on \ac{mIoU}. This rapid, low-data convergence enables on-device adaptation: from a 5-minute recording of the SNOW-2 Grand Tour sequence~\citep{frey2026grandtour} we produced a deployed model in under 15 minutes, with data curation reducing 3{,}900 frames to 20 images in 2 minutes, SAM3-assisted 3-class text-prompt labeling in under 5 minutes, and on-device training plus TensorRT export in 4 minutes.

We applied the same on-device workflow to two further Grand Tour sequences, ETH-2 (an indoor building) and ICE-1 (an ice cave), yielding three sites with 20 clustered, SAM3-labeled images each, split into 16 training and 4 validation images (Table~\ref{tab:grandtour-adaptation}). Each site uses a coarse three-class layer comprising a traversable surface class, a non-traversable class, and a background class that absorbs all unprompted pixels. SAM3 is prompted with a single noun phrase per non-background class; SNOW-2 uses the prompts ``snowy floor'' and ``fence'', ETH-2 uses ``indoor floor'' and ``people'', and ICE-1 uses ``ice floor'' and ``glass barrier'', reconstructed after the fact and matching the original labels at 0.94~\ac{mIoU}. The resulting masks are reviewed by hand before training. All decoders are trained on the AGX Orin with early stopping on validation \ac{mIoU}, and the validation frames are the held-out cluster representatives of the same recording. ICE-1 yields the lowest score of the three, illustrating a site at which further samples would be added through the same workflow.

\begin{table}[t]
\centering
\caption{\textbf{Rapid adaptation on three Grand Tour sites.} Best validation \ac{mIoU} and pixel accuracy (3 classes) of the decoder trained on the AGX Orin from 16 clustered, SAM3-labeled images and evaluated on the 4 held-out cluster representatives of the same recording, with early stopping on validation \ac{mIoU}. Training time is the wall-clock time of the training step alone on the AGX Orin, excluding data curation, SAM3 labeling, and TensorRT export. Single run per site.}
\label{tab:grandtour-adaptation}
\small
\setlength{\tabcolsep}{4pt}
\begin{tabular}{llccc}
\hline
Sequence & Scene & \ac{mIoU} [\%] & Acc [\%] & \shortstack[c]{Training\\time [s]} \\
\hline
SNOW-2 & alpine snow field & 90.7 & 97.4 & 177 \\
ETH-2 & indoor building & 89.0 & 97.1 & 90 \\
ICE-1 & ice cave & 77.3 & 88.8 & 123 \\
\hline
\end{tabular}
\end{table}

On our outdoor trail dataset, we additionally quantify the impact of the label budget, training the decoder on subsets of $k$ images selected with the same clustering pipeline (mean $\pm$ SD over $n=3$ seeds). With 20 labeled images, the decoder reaches $82.6 \pm 0.1\%$ \ac{mIoU} on the 3{,}196-image test split; the full 28{,}763-image training set raises this to $90.0\%$ ($97.0\%$ pixel accuracy), with returns diminishing smoothly in between (Table~\ref{tab:trail-label-efficiency}). Convergence mirrors the RELLIS-3D study: for $k \geq 100$, validation \ac{mIoU} is within $1\%$ of its final value by the second epoch. Baseline comparisons are reported on the public RELLIS-3D benchmark above; here we characterize the label-budget trade-off on our deployment domain.

\begin{table}[t]
\centering
\caption{\textbf{Label efficiency on the outdoor trail dataset.} Best validation \ac{mIoU} on the 3{,}196-image test split versus the number of labeled training images $k$ (mean $\pm$ SD over $n=3$ seeds).}
\label{tab:trail-label-efficiency}
\small
\begin{tabular}{r c}
\hline
$k$ & \ac{mIoU} [\%] \\
\hline
20 & $82.61 \pm 0.09$ \\
50 & $83.10 \pm 0.06$ \\
100 & $83.80 \pm 0.04$ \\
200 & $84.23 \pm 0.05$ \\
500 & $84.83 \pm 0.07$ \\
1{,}000 & $85.84 \pm 0.30$ \\
2{,}000 & $87.22 \pm 0.21$ \\
4{,}000 & $88.18 \pm 0.23$ \\
28{,}763 (full) & $90.01 \pm 0.08$ \\
\hline
\end{tabular}
\end{table}

\section{Confidence-Weighted Map Fusion} \label{supp:fusion}
Each grid cell is updated by a confidence-weighted fusion of the projected traversability observations. For a cell with a valid image correspondence and per-pixel confidence $c_p \ge c_{\text{floor}}$, we form a distance weight
\begin{equation}
    w_d = \frac{1}{1 + (d/r)^4},
\end{equation}
where $d$ is the ground-plane distance from the camera to the cell and $r$ is the fusion range (3~m in the indoor deployment; by default a quarter of the map extent). The observation confidence is $c = c_p\, w_d$, and the cell value $v$ is blended toward the observation $o$ as
\begin{equation}
    v \leftarrow v + \alpha\,(o - v), \qquad \alpha = \frac{c}{c + W},
\end{equation}
where $W$ is the cell's accumulated evidence, initialized to $0$ so that the first valid observation is written directly ($\alpha = 1$). The evidence is then updated as $W \leftarrow \min(W + c,\, W_{\max})$; the cap $W_{\max}$ bounds how confident a cell becomes, preserving plasticity to later change. Cells without a valid correspondence, or with $c_p < c_{\text{floor}}$, retain their value and evidence, and the evidence layer is shifted and cleared together with the map during recentering. We use $c_{\text{floor}} = 0.5$ and $W_{\max} = 4.0$.

\section{Confidence and Out-of-Distribution Detection} \label{supp:ood}
To test whether the confidence signal detects unseen obstacles, we train all architectures on Cityscapes~\citep{cordts2016cityscapes} using 200 of 2{,}975 images and evaluate on the Lost and Found~\citep{pinggera2016lostandfound} test set, which contains real road obstacles absent from training. We report the object-centric false-positive rate (obj-FPR), the fraction of obstacle pixels misclassified as traversable. \ac{SemMap} uses cosine-similarity confidence; baselines use maximum softmax probability.

\begin{figure}[tbp]
\centering
\includegraphics{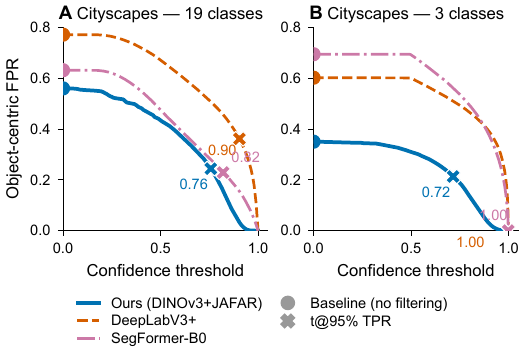}
\caption{\textbf{\acs{OOD} obstacle detection on Lost and Found} (trained on Cityscapes with 200 images; evaluated on the $1{,}111$ test images containing obstacles). Pooled object-centric FPR vs.\ confidence threshold at 20 and 3 classes. Markers: no filtering (circle), $95\%$ TPR threshold (triangle).}
\label{fig:ood-study}
\end{figure}

\begin{figure*}[h!]
\centering
\includegraphics[width=\textwidth]{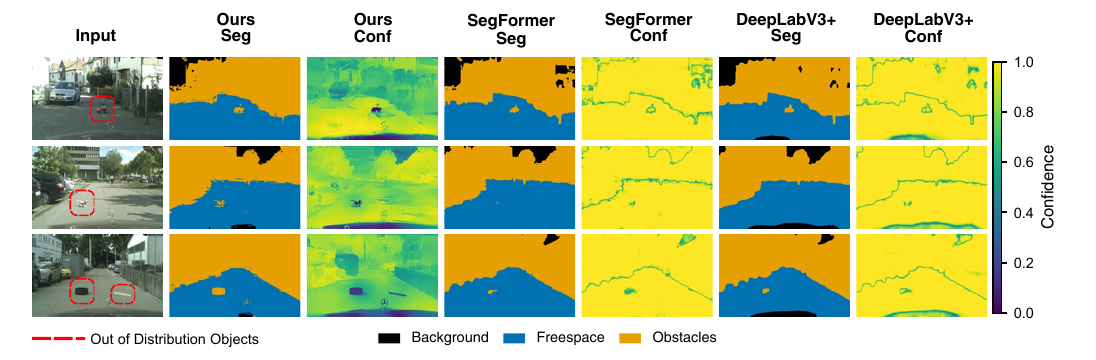}
\caption{\textbf{Qualitative \acs{OOD} detection on Lost and Found.} For each method: segmentation overlay and confidence map. Row 1: all methods detect the obstacle. Row 2: baselines fail while their softmax confidence stays high. Row 3: SemMap misclassifies the right obstacle, but cosine-similarity confidence is clearly low there, enabling recovery.}
\label{fig:lostnfound-conf}
\end{figure*}

Without filtering, \ac{SemMap} predicts $1.7$--$2.0\times$ fewer obstacle pixels as traversable than the baselines at 3 classes: $35.0\%$ of obstacle pixels versus $60.2\%$ for DeepLabV3+ and $69.4\%$ for SegFormer-B0, pooled across the $1{,}111$-image test set (Fig.~\ref{fig:ood-study}). The per-image difference is significant (paired Wilcoxon signed-rank, $P<0.001$ Holm-corrected; rank-biserial $r=0.76$ and $0.85$ against DeepLabV3+ and SegFormer-B0, respectively). At 19 classes all methods degrade, but \ac{SemMap} ($56\%$) still beats the baselines ($63$--$77\%$). Reducing from 19 to 3 classes helps \ac{SemMap} most ($56\%\rightarrow35\%$) while SegFormer-B0 worsens ($63\%\rightarrow69\%$), indicating that a coarser, traversability-focused label space suits cosine-similarity confidence. At a $95\%$ true-positive rate, both baselines require threshold $t=1.0$ at 3 classes, leaving no tunable range, whereas \ac{SemMap} reaches it at $t=0.72$ with a wide usable range. This is not a substitute for explicit synthetic-negative training, which could improve \ac{OOD} detection further; the advantage is meaningful \ac{OOD} awareness at negligible additional training and inference cost. Fig.~\ref{fig:lostnfound-conf} shows the qualitative behavior: softmax confidence highlights only class boundaries, while cosine similarity is clearly low at misclassified obstacles, enabling recovery through downstream filtering. The same confidence signal drives the on-robot elevation-map fusion during indoor racing (Fig.~\ref{fig:indoor}D): with confidence-weighted fusion enabled (Appendix section~\ref{supp:fusion}) the accumulated semantic map stays consistent across laps, whereas without it the low-confidence predictions from high-speed turns progressively corrupt the map.

\section{Inference Performance} \label{supp:upsampler}
We profile \ac{SemMap} on an NVIDIA AGX Orin 64GB (MAXN, max fan, CUDA \ac{MPS}, FP16, $320\times512$ input, 50 warmup and 100 timed iterations) and measure segmentation quality on the ORFD~\citep{min2022orfd} test set (2{,}193 images) with the decoder trained on 500 curated images. Each timed iteration passes the same pre-loaded image through the network, so the measurements capture pure inference time and exclude message transport and synchronization overheads; the pipeline is also profiled in isolation, without the rest of the autonomy stack running. The reported throughputs are therefore upper bounds, which is why we state them as ``up to'' throughout.

The configurations in Table~\ref{tab:inference} time model inference including confidence, i.e., backbone, upsampler, segmentation head, and cosine-similarity confidence. To obtain the complete per-frame cost, we additionally time the call that the semantic worker executes for every frame, with the clocks pinned (\texttt{jetson\_clocks}) and TensorRT~10.7. This call adds input preprocessing, the temperature-weighted traversability of Eq.~\eqref{eq:traversability}, and the upsampling of segmentation, traversability, and confidence to input resolution. It excludes the host-to-device image upload, since the worker receives frames already resident in GPU memory. Averaged over three runs with two different segmentation layers, the complete semantic pipeline takes \SI{11.12}{\milli\second} (\SI{90}{\hertz}; run means between \SI{10.93}{\milli\second} and \SI{11.30}{\milli\second}) with JAFAR$\times4$ and \SI{6.12}{\milli\second} (\SI{163}{\hertz}) with bilinear$\times4$ upsampling, of which the upsampler accounts for \SI{4.90}{\milli\second} and \SI{0.26}{\milli\second}, respectively. The number of classes in the segmentation layer has no measurable effect on these timings.

\begin{figure}[tbp]
\centering
\includegraphics{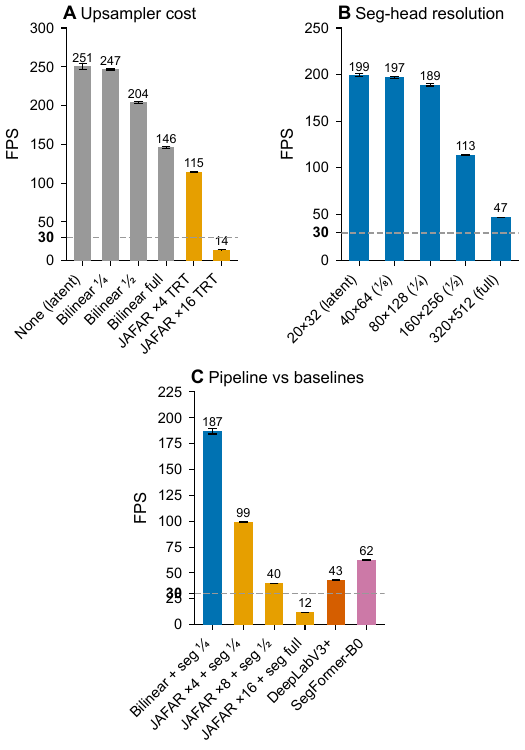}
\caption{\textbf{Inference profiling on AGX Orin 64GB.} \textbf{(A)} Upsampler cost: bilinear is nearly free while JAFAR scales with the upsampling factor. \textbf{(B)} Segmentation-head resolution with bilinear upsampling. \textbf{(C)} End-to-end pipeline configurations (all TensorRT).}
\label{fig:inference-study}
\end{figure}

Converting the DINOv3-s backbone from PyTorch to TensorRT gives an approximately $4\times$ speedup ($56\rightarrow228$~\acs{FPS}) with no quality change (Table~\ref{tab:inference}, Fig.~\ref{fig:inference-study}). Bilinear upsampling yields identical \ac{mIoU} ($83.8\%$) at any resolution, confirming it adds no new information; since interpolation only resizes the predictions, this value also serves as the no-upsampler baseline, the quality of the raw patch-level predictions. JAFAR$\times4$ at quarter resolution improves \ac{mIoU} by $4.1$ percentage points ($87.9\%$) at~\SI{99}{\fps} by attending to image-level cues, our recommended setting; $\times8$ and $\times16$ add under $0.2$ percentage points while dropping to 40 and \SI{12}{\fps}. This per-image gain is statistically robust: on the ORFD test set ($n=2{,}193$ images), JAFAR improves per-image \ac{mIoU} by $+3.5$ to $+3.8$ points over bilinear at matched resolution (paired Wilcoxon signed-rank, $P<0.001$ Holm-corrected; rank-biserial $r=0.92$--$0.97$; JAFAR wins on $89$--$94\%$ of images). Our recommended configuration is also faster than the fine-tuned segmentation baselines profiled identically on the AGX Orin:~\SI{99}{\fps} versus \SI{43}{\fps} for DeepLabV3+ and \SI{62}{\fps} for SegFormer-B0 ($1.6$--$2.3\times$). Unlike self-supervised methods~\citep{mattamala2025wvn, jung2024vstrong, aegidius2025stepp} that incur online-learning overhead at deployment, \ac{SemMap} needs only a single forward pass through a frozen backbone.

\begin{table}[t]
\centering
\caption{\textbf{Inference latency on AGX Orin 64GB} ($320\times512$) and segmentation quality on ORFD. Latencies are the mean over $n=100$ timed iterations (50 warm-up); per-configuration SD $\le 0.33$~ms. TensorRT FP16 unless noted. All rows above the last block time model inference including confidence; the last block times the complete semantic pipeline per frame, adding preprocessing, traversability, and output upsampling, averaged over repeated runs (see text). Arrows denote the preferred direction ($\downarrow$: lower is better, $\uparrow$: higher is better). The bilinear configurations equal the no-upsampler baseline, as interpolation adds no information. $^\star$Recommended configuration.}
\label{tab:inference}
\setlength{\tabcolsep}{3pt}
\begin{tabular}{lcccc}
\hline
Configuration & ms $\downarrow$ & \acs{FPS} $\uparrow$ & \acs{mIoU} $\uparrow$ & Acc $\uparrow$ \\
\hline
Backbone only (PyTorch) & 17.74 & 56 & -- & -- \\
Backbone only (TRT) & 4.38 & 228 & -- & -- \\
\hline
Bilinear + seg 1/4 & 5.34 & 187 & 83.81 & 93.73 \\
Bilinear + seg 1/2 & 8.82 & 113 & 83.85 & 93.75 \\
Bilinear + seg full & 21.42 & 47 & 83.86 & 93.75 \\
JAFAR$\times4$ + seg 1/4$^\star$ & 10.11 & 99 & 87.90 & 95.19 \\
JAFAR$\times8$ + seg 1/2 & 25.13 & 40 & 88.05 & 95.25 \\
JAFAR$\times16$ + seg full & 84.78 & 12 & 87.91 & 95.17 \\
\hline
Full pipeline, bilinear$\times4$ & 6.12 & 163 & -- & -- \\
Full pipeline, JAFAR$\times4$ & 11.12 & 90 & -- & -- \\
\hline
\end{tabular}
\end{table}

\subsection{System Dataflow} \label{supp:dataflow}
\begin{figure}[tbp]
\centering
\includegraphics{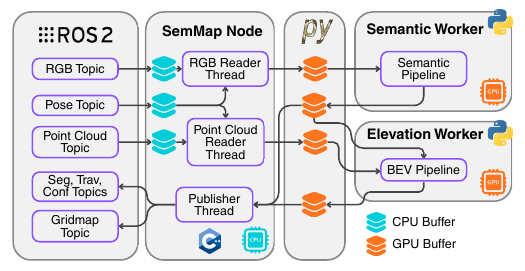}
\caption{\textbf{Dataflow architecture.} A C++ ROS2 node subscribes to RGB, pose, and point cloud topics via dedicated reader threads, managing CPU buffers for synchronization. Data is passed to the Python layer via pybind11 into shared GPU buffers, which feed the semantic and elevation-mapping worker processes running as separate Python processes. Outputs (traversability, segmentation, confidence, grid map) are published back through a publisher thread. All inter-process communication uses GPU buffers to avoid memory transfers.}
\label{fig:dataflow}
\end{figure}

Fig.~\ref{fig:dataflow} details the hybrid C++/Python dataflow of \ac{SemMap} described in the main text, which lets the high-rate input streams be consumed without \ac{GIL} contention. The reader threads hold the RGB and point-cloud streams in double buffers and poses in a ring buffer of 1{,}000 entries; buffer sizes follow the topic rates, with the pose buffer sized by a rule of thumb of one second of the~1000~Hz pose stream. Because sensor messages are held until a matching pose is available for interpolation, a lagging pose source, as we observed when testing with ZED visual-inertial odometry, requires correspondingly larger RGB and point-cloud buffers. CUDA \ac{MPS} lets the two workers execute kernels concurrently, improving GPU utilization.

\section{Cross-Embodiment Deployment on a Humanoid} \label{supp:g1}
\begin{figure}[tbp]
\centering
\includegraphics{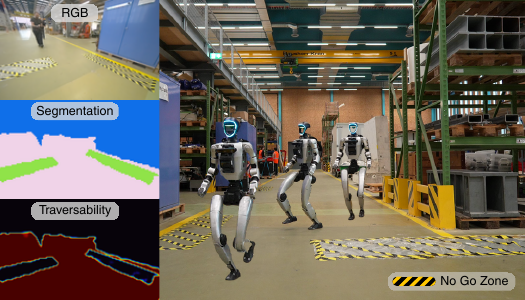}
\caption{\textbf{Cross-embodiment deployment on a Unitree G1 humanoid.} SemMap runs in RGB-only mode and segments taped no-go zones into a 2D traversability image for the navigation policy. Only the decoder is adapted; the frozen backbone and upsampler are shared with the bicycle deployment.}
\label{fig:teaser-g1}
\end{figure}

To show that \ac{SemMap} is not specific to the bicycle, we deploy it on a Unitree G1 humanoid for factory navigation. The pipeline runs in semantic-only mode from a single ZEDX RGB camera, producing a 3-class segmentation that is converted to a 2D traversability image and consumed by an \ac{RL} navigation policy. The task is no-go-zone avoidance: restricted areas demarcated by tape, where the enclosed region must be classified as non-traversable from its boundary rather than its surface (Fig.~\ref{fig:teaser-g1}). The same frozen backbone and upsampler used on the bicycle are reused unchanged, with only the decoder and class-to-traversability mapping adapted to the factory classes, demonstrating cross-embodiment transfer between a high-speed wheeled robot consuming semantic elevation maps and a legged humanoid consuming 2D traversability images. As with our indoor and outdoor trail datasets, the factory training images were captured primarily with a handheld capture device and, in some cases, a smartphone, decoupling data collection from deployment.

\section{Field-Deployment Summary} \label{supp:deployment_days}
Our field deployment on the Allmend trail progressed through four testing days, summarized below:
\begin{itemize}
    \item \textbf{April 8:} Initial autonomous tests comprising 24 runs. The system traversed long segments of the lower Allmend trail, reaching distances of up to \SI{242}{\meter} (\SI{76}{\second}), and leading to a median traveled distance of \SI{77}{\meter}. However, the policy lacked sufficient robustness to handle the human-made obstacles on the upper part of the trail, losing balance shortly after encountering them, exemplified by a median survival distance of \SI{25}{\meter}.
    \item \textbf{April 24:} Testing was stopped after 10 runs (8 on the lower and 2 on the upper trail), with median traveled distances of \SI{25}{\meter} and \SI{6}{\meter}, respectively. Between the two April deployments, the trees along the trail had grown new leaves, and trees and grasses were flowering, changing the trail's appearance with fresh foliage, patchy shade, flowers and pollen, and less of the trail was segmented as traversable: compared to April 8th, a larger share of the driven path was marked untraversable (\SI{12}{\percent} vs.\ \SI{3}{\percent}). In addition, without semantic inpainting, trail regions that the camera had not labelled remained untraversable. Before the next deployment, we recorded new footage on the trail, retrained the decoder, and introduced joint semantic and geometric inpainting.
    \item \textbf{May 8:} Autonomous navigation across 24 runs on both the upper and lower trail segments, with a longest continuous traversal of \SI{233}{\meter} (\SI{74}{\second}). The median on both the upper and lower parts was quite similar (\SI{90}{\meter} and \SI{74}{\meter}). This deployment was the first to combine semantic-based inpainting with a version of \ac{SemMap} trained on an expanded dataset, and it determined the tuning parameters used on the final day.
    \item \textbf{May 28:} Autonomous deployment comprising 21 runs, with continuous traversals of up to \SI{210}{\meter} (\SI{51}{\second}). For this day, we slightly widened the domain randomization ranges; all randomization parameters reported in this work refer to this final policy.
\end{itemize}
On each testing day, field operations began at approximately 10:00 and ended between 14:00 and 17:00, depending on system performance and data collection requirements.

\begin{figure}[h]
    \centering
    \includegraphics[width=\columnwidth]{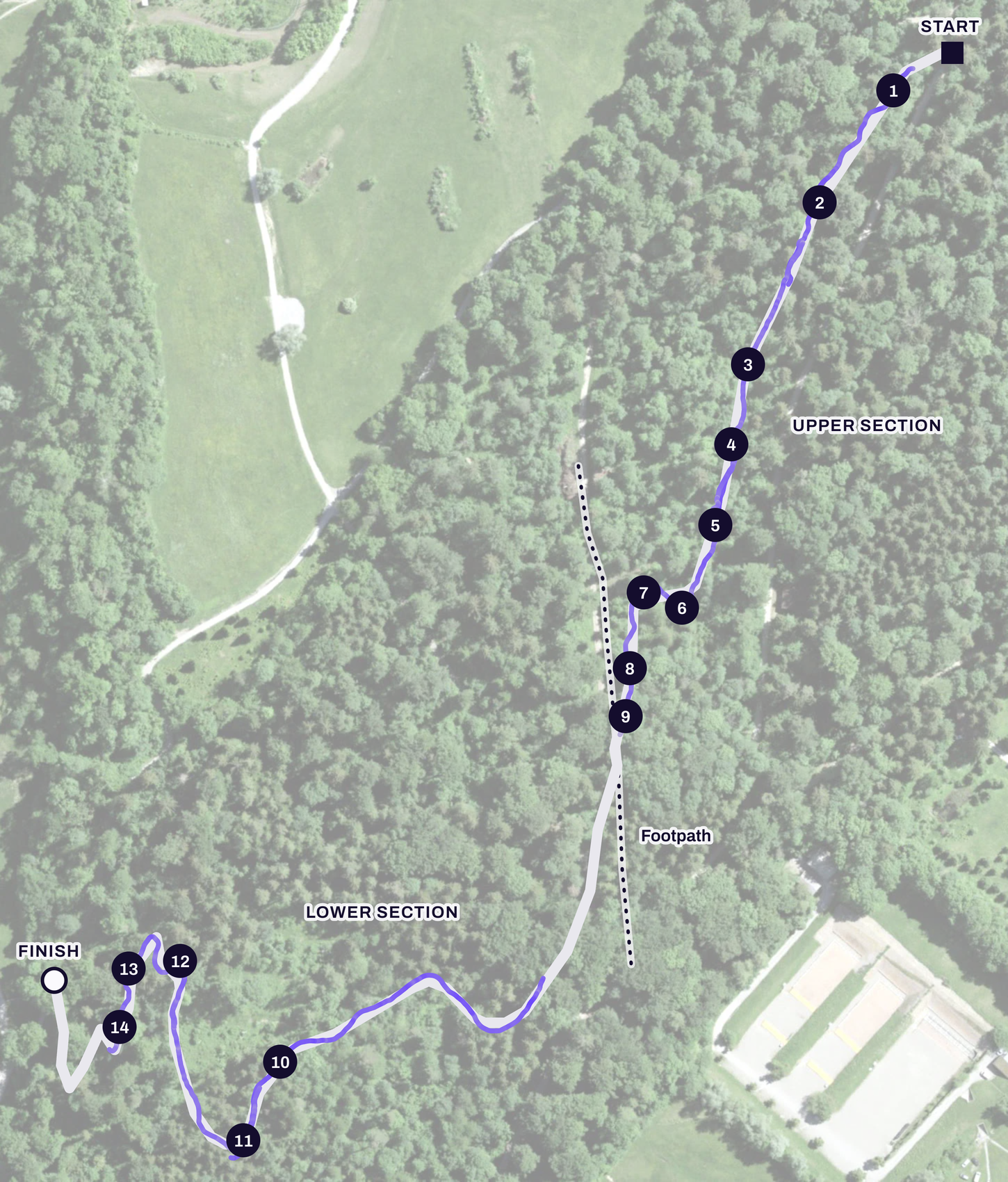}
    \caption{\textbf{Trail coverage.} Map of the Allmend trail (light grey) overlaid with the estimated base trajectories of the six runs in Table~\ref{tab:allmend_real} (purple).}
    \label{fig:trail_coverage}
\end{figure}

Figure~\ref{fig:trail_coverage} shows the upper and lower segments of the Allmend trail with the checkpoints of Fig.~\ref{fig:trail_racing}, overlaid with the base trajectories of the six runs reported in Table~\ref{tab:allmend_real}, as estimated by our onboard \ac{LIO} state estimator.

\begin{figure*}
    \centering
    \includegraphics[width=\textwidth]{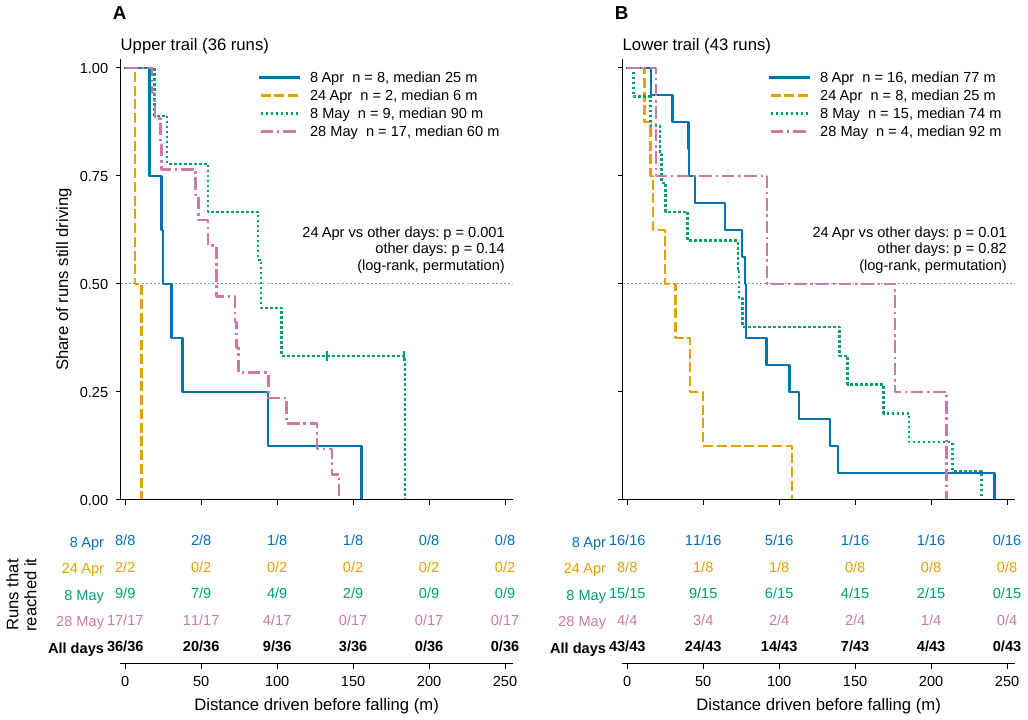}
    \caption{\textbf{Deployment statistics.} Kaplan–Meier estimates of the distance driven before a fall for all four deployment days (79 runs), shown separately for the upper (left, 36 runs) and lower (right, 43 runs) parts of the Allmend trail. Two of the April 24 runs were recorded on the upper part. The dotted line at 0.5 marks the median: the distance by which half of a day's runs had fallen (values in the legend). Plus marks denote the two runs that ended in a stable track-stand.}
    \label{fig:deployment_survival}
\end{figure*}

To evaluate system robustness, we analyzed the distance driven before a fall across all 79 runs using the Kaplan–Meier method (Fig.~\ref{fig:deployment_survival}), with 36 runs on the upper and 43 runs on the lower part of the trail, and two runs ending in a stable track-stand. On the three evaluated days, the distance before a fall did not differ significantly between days on either trail section; only the April 24 deployment differs significantly. Runs on the lower trail typically covered more distance than runs on the upper trail, reflecting the difficulty of the human-made obstacles on the upper part.

\section{Ablation and Evaluation of Racing Policies}
The evaluation environment comprised 10 instances of each terrain type across five difficulty levels, yielding a total of 850 distinct terrain patches.  We simulated 4{,}000 robots per difficulty level with all domain randomizations disabled, and recorded the survival rate, time to reach the final platform, time spent off-trail, and average velocity.

\begin{table*}[htbp]
\centering
\caption{\textbf{Comparison of policies in simulation.} Policy-1: trained with termination upon leaving the trail. Policy-2: trained without termination upon leaving the trail. Policy-3: trained without domain randomization. Each experiment (row) is averaged over 4{,}000 agents, with a single fixed seed. Training involved 3{,}000 learning iterations and we set $v_\text{min}=1$.} 
\label{tab:policy_evaluation}
\begin{tabular}{@{}lccccc@{}}
\toprule
\textbf{Policy} & \textbf{Level} & \textbf{Survival Rate} & \textbf{Time to Goal} & \textbf{Time Not on Trail} & \textbf{Avg Speed} \\ 
& & & (s) & (s) & (m/s) \\ \midrule
policy-1 & 0 & 0.91 & 11.3 & 0.023 & 2.58 \\
 & 1 & 0.90 & 11.3 & 0.022 & 2.59 \\
 & 2 & 0.88 & 11.5 & 0.020 & 2.58 \\
 & 3 & 0.82 & 11.8 & 0.021 & 2.58 \\
 & 4 & 0.70 & 12.3 & 0.057 & 2.55 \\ \midrule
policy-2 & 0 & 0.94 & 10.6 & 0.017 & 2.74 \\
 & 1 & 0.93 & 10.8 & 0.030 & 2.70 \\
 & 2 & 0.92 & 10.9 & 0.018 & 2.72 \\
 & 3 & 0.86 & 11.2 & 0.015 & 2.71 \\
 & 4 & 0.78 & 11.8 & 0.051 & 2.67 \\ \midrule
policy-3 & 0 & 0.95 & 12.0 & 0.014 & 2.41 \\
 & 1 & 0.97 & 11.3 & 0.019 & 2.58 \\
 & 2 & 0.97 & 11.1 & 0.017 & 2.65 \\
 & 3 & 0.94 & 11.1 & 0.017 & 2.70 \\
 & 4 & 0.88 & 11.6 & 0.044 & 2.68 \\ \bottomrule
\end{tabular}
\end{table*}

\subsection{Hard Termination vs. Soft Reward} \label{supp:track_following}
We compared two racing policies trained using the \ac{TL} for 3{,}000 learning iterations, and $v_\text{min}=1$ (for faster convergence) under different boundary enforcement strategies: a hard termination upon trail violation (policy-1) versus a soft continuous reward for trail following (policy-2). 

We evaluated the final trained policies in simulation to benchmark their comparative performance (Tab.~\ref{tab:policy_evaluation}, policy-1 and policy-2). In both cases, we did not terminate upon trail boundary violation. The data reveals that both policies adhere to the trail boundary constraints equally well, and policy-2 (average speed \SI{2.726}{\meter\per\second}) is slightly faster and more robust than policy-1 (average speed \SI{2.576}{\meter\per\second}).


\subsection{With and without domain Randomization} \label{supp:track_rand}
To isolate the impact of domain randomization, we trained an ablation policy with all randomization events and noise terms disabled during both training and evaluation. We use the same training parameters as before (3{,}000 learning, $v_\text{min}=1$ ) for consistent comparison. As detailed in Table~\ref{tab:policy_evaluation} as policy-3, the absence of domain randomization leads to slightly lower average speed (\SI{2.6}{\meter\per\second}) at comparable survival rates. We note that policies trained without these randomizations fail to transfer to the physical robot (strong vibrations lead to instantaneous falls).

\end{appendix}

\end{document}